\documentclass[11pt]{article}

\usepackage[preprint]{acl}

\usepackage{times}
\usepackage{latexsym}

\usepackage[T1]{fontenc}

\usepackage[utf8]{inputenc}

\usepackage{microtype}

\usepackage{inconsolata}

\usepackage{graphicx}

\usepackage{booktabs}
\usepackage{colortbl}
\usepackage[table]{xcolor}
\usepackage{array}
\usepackage{makecell}
\usepackage{graphicx}
\usepackage{float}
\usepackage{amsmath}
\usepackage{amssymb}

\title{Dynamic Commonsense Coordination for \\ Empathetic Response Generation}

\author{
Zhengyu Qi \\
Leiden Institute of Advanced Computer Science (LIACS) \\
Leiden University, The Netherlands \\
\texttt{qiwendell@gmail.com}
}

\begin{document}
\maketitle
\begin{abstract}

Empathetic Response Generation (ERG) requires models to recognize users' emotions and generate empathetic responses. 
Commonsense knowledge has been shown to support such reasoning, yet existing approaches typically reuse fixed commonsense representations across understanding and generation, limiting their ability to coordinate such knowledge across different stages.
We propose DCC, a Dynamic Commonsense Coordination Framework with three complementary modules: residual-based commonsense interaction (SCE-AttnRes) to integrate contextual and situational commonsense representations, Association-Guided Commonsense Filtering (AGCF) to down-weight low-relevance commonsense relations, and Iterative Commonsense-Aware Decoding (ICAD) to dynamically retrieve commonsense memories during generation.
Experiments on the Empathetic-Dialogues benchmark show that DCC improves emotion classification accuracy and response diversity over the CEM baseline while maintaining comparable perplexity.
An LLM-based blind evaluation further demonstrates that DCC generates responses with better relevance, coherence, and informativeness.
The code and implementation details will be publicly available at \url{https://github.com/Hanabi-Q/DCC-ERG}.

\end{abstract}

\section{Introduction}

Empathetic Response Generation (ERG) has become an important research direction in open-domain dialogue systems. 
Unlike general dialogue generation, ERG requires models to understand users’ implicit emotional states and generate emotionally appropriate responses. 
Rashkin et al.~\cite{DBLP:conf/acl/RashkinSLB19} emphasize that empathetic systems should recognize and respond to speakers’ implied feelings.

Nevertheless, in real-world dialogues, recognizing emotions alone is often not sufficient. 
The emotions expressed by users are usually closely related to specific events, situations, and underlying psychological states. 
For example, anxiety may stem from work pressure, while feelings of disappointment may be caused by interpersonal conflicts. 
Therefore, an effective empathetic response should not only perceive the emotion itself, but also understand the underlying situations and causes behind the emotion. 
This view is consistent with psychological studies suggesting that empathy involves both cognitive and affective components, shaped by contextual appraisal and interpersonal context~\cite{davis1983measuring,smith2006cognitive,singer2009social}.
These findings highlight the importance of modeling users’ emotional and cognitive states in ERG.

Recent ERG approaches have therefore introduced commonsense knowledge to improve the understanding of users' emotional and cognitive states. 
Building upon this idea, subsequent studies have further explored commonsense reasoning~\cite{DBLP:conf/aaai/SabourZH22}, situation-aware interaction~\cite{DBLP:journals/ipm/YangRWSZL24}, dynamic emotion-semantic modeling~\cite{DBLP:conf/emnlp/YangRYZCCWSJL23}, and iterative associative reasoning~\cite{DBLP:conf/acl/YangRYSCZL24} to enhance empathetic dialogue generation.

Despite these advances, commonsense representations are typically constructed once and reused across both dialogue understanding and response generation, with limited adaptation to different interaction stages.

In practice, the role of commonsense may change as dialogue understanding gradually shifts to response generation.
Without adaptive commonsense coordination, irrelevant or low-relevance commonsense information may introduce noisy reasoning and negatively affect response quality and empathetic consistency~\cite{DBLP:conf/acl/CaiSXSWGZX23}.

Motivated by these observations, we propose a Dynamic Commonsense Coordination Framework for commonsense-aware empathetic dialogue generation. 
Specifically, the framework consists of three complementary components: residual-based~\cite{DBLP:journals/corr/abs-2603-15031} commonsense interaction (Situation Commonsense Enhancement, SCE-AttnRes) to adaptively integrate contextual and situational commonsense, Association-Guided Commonsense Filtering (AGCF) to filter low-relevance commonsense relations, and Iterative Commonsense-Aware Decoding (ICAD) to dynamically utilize commonsense during response generation.

The main contributions of this work are summarized as follows:

\begin{itemize}
\item We propose a Dynamic Commonsense Coordination Framework that addresses the limitations of static commonsense utilization by dynamically coordinating commonsense across dialogue understanding and response generation.

\item We introduce three complementary modules: SCE-AttnRes for cross-source integration, AGCF for relation-level filtering, and ICAD for token-level retrieval, which together enable stage-aware commonsense coordination.

\item Extensive experiments on Empathetic-Dialogues benchmark~\cite{DBLP:conf/acl/RashkinSLB19} demonstrate substantial improvements over the underlying CEM~\cite{DBLP:conf/aaai/SabourZH22} baseline, while ablation and mechanism analyses clarify the distinct roles of the proposed components.

\end{itemize}

\section{Related Work}
\subsection{Emotion-aware Empathetic Response Generation} 
Early studies on empathetic dialogue generation primarily focused on modeling users' emotions to generate empathetic responses.
Rashkin et al.~\cite{DBLP:conf/acl/RashkinSLB19} introduced the EMPATHETIC-DIALOGUES dataset and established empathetic dialogue generation as an important task in open-domain dialogue systems.
Following this setting, MoEL~\cite{DBLP:conf/emnlp/LinMSXF19}, MIME~\cite{DBLP:conf/emnlp/MajumderHPLGGMP20}, and EmpDG~\cite{DBLP:conf/coling/LiCRRTC20} improved empathetic dialogue generation through emotion-specific experts, emotion mimicry, and multi-resolution emotion modeling.

These methods mainly focus on modeling emotional signals from the dialogue context, while commonsense reasoning about users' situations and cognitive states is not explicitly considered. 
This motivates subsequent studies to incorporate external commonsense knowledge into empathetic dialogue generation.

\subsection{Commonsense-aware Empathetic Response Generation}

To improve the understanding of users’ emotional and cognitive states, increasing efforts have introduced commonsense knowledge into empathetic dialogue generation. KEMP~\cite{DBLP:conf/aaai/LiLRRC22} first incorporated ConceptNet~\cite{DBLP:conf/aaai/SpeerCH17} and NRC-VAD emotional lexicon\cite{DBLP:conf/acl/Mohammad18} to enhance the understanding of implicit emotional information. 
Later, CEM~\cite{DBLP:conf/aaai/SabourZH22} explicitly modeled users' cognitive and affective states by leveraging COMET-generated~\cite{DBLP:conf/acl/BosselutRSMCC19,DBLP:conf/aaai/HwangBBDSBC21} commonsense knowledge to infer implicit intents, needs, and reactions.

Subsequent studies further explored more complex commonsense interaction patterns and dynamic empathy modeling.  
SEEK~\cite{DBLP:conf/emnlp/WangLLMYWZ22} employed serial encoding and bidirectional knowledge selection to capture emotion transitions across dialogue turns.
CASE~\cite{DBLP:conf/acl/ZhouZW0H23} aligned cognition and affection at both coarse- and fine-grained levels to enhance empathetic reasoning.
SDAM~\cite{DBLP:journals/ipm/YangRWSZL24} emphasized the importance of situation-dialogue association for empathy understanding. IAMM~\cite{DBLP:conf/acl/YangRYSCZL24} further modeled fine-grained associations among dialogue utterances, situation information, and commonsense knowledge through iterative associative memory interaction.

Despite these advances, existing commonsense-aware ERG methods generally lack dynamic coordination of commonsense utilization across dialogue understanding and response generation. 
Commonsense representations are typically constructed before generation and reused across stages, making it difficult to adapt their utilization to the distinct requirements of different interaction stages.

Without adaptive coordination, irrelevant or low-relevance commonsense information may introduce noisy reasoning and reduce response quality and empathetic consistency.

\subsection{Dynamic Modeling and Reasoning for Empathetic Response Generation}

Another line of research has focused on dynamic emotion modeling and interaction mechanisms for empathetic response generation.
ESCM~\cite{DBLP:conf/emnlp/YangRYZCCWSJL23} explored dynamic emotion-semantic correlation to better model the interaction between emotional and semantic information during empathetic response generation.
E-CORE~\cite{DBLP:conf/emnlp/FuZWM23} explicitly modeled intrinsic emotion correlation to enhance both emotion perception and response generation across the two stages.

Alongside these emotion-centric dynamic modeling approaches, recent studies have explored more explicit reasoning mechanisms for empathetic response generation. 
CFEG~\cite{DBLP:journals/corr/abs-2408-11599} incorporates emotion-cause reasoning and COMET-generated commonsense into chain-of-thought fine-tuning for large language models.
EICL~\cite{DBLP:conf/cikm/Ren0YSCZL25} enhances fine-grained emotion reasoning and decision-making through emotionally-aware in-context learning. 
ReflectDiffu~\cite{DBLP:conf/acl/YuanDCYN25} bridges emotion and intent via a reflection mechanism grounded in emotional contagion and empathetic mimicry. 
STRIDE-ED~\cite{DBLP:journals/corr/abs-2604-07100} further formulates empathetic dialogue generation as a strategy-grounded, multi-stage reasoning process with explicit strategy planning and stepwise decision-making.

These studies make empathetic reasoning more explicit and interpretable by introducing structured reasoning, reflection, or strategy planning. Nevertheless, the dynamic coordination of commonsense utilization across dialogue understanding and response generation remains largely underexplored.

\section{Preliminaries}

\subsection{Task Formulation}
\begin{figure}[t]
    \centering
    \includegraphics[width=\columnwidth]{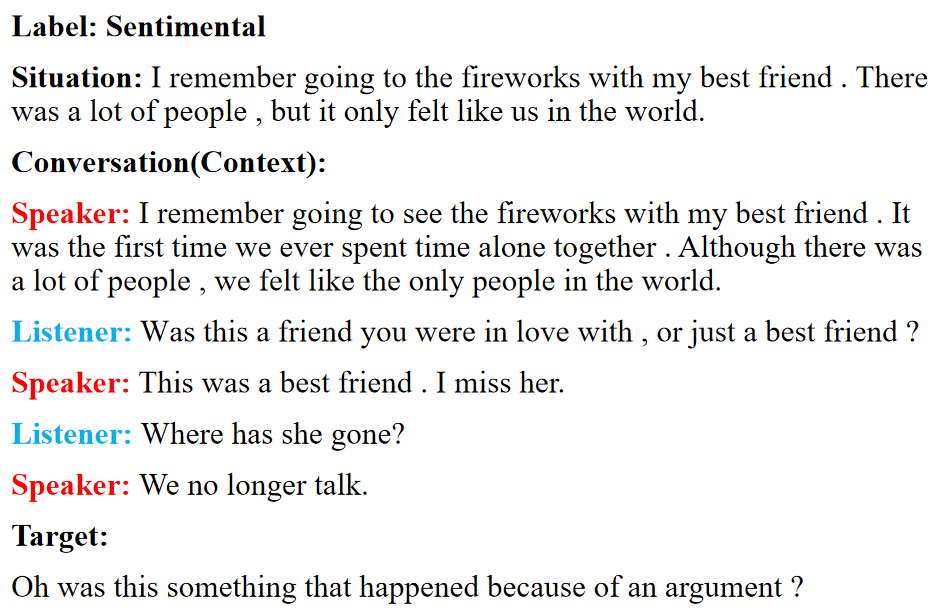}
    \caption{An example from the Empathetic-Dialogues dataset. Each sample contains an emotion label, a situation description, a multi-turn dialogue context, and a target empathetic response.}
    \label{fig:example}
\end{figure}

Figure~\ref{fig:example} presents an example from the Empathetic-Dialogues dataset~\cite{DBLP:conf/acl/RashkinSLB19}. 
Each sample contains an emotion label, a situation description providing the dialogue background, a multi-turn dialogue context between a speaker and a listener, and a target response. 
The goal is to infer the speaker's emotional state from the dialogue context and situation description, then generate an appropriate empathetic response.

Following recent work~\cite{DBLP:journals/ipm/YangRWSZL24, DBLP:conf/acl/YangRYSCZL24}, we adopt the same situation-aware setting where the model jointly performs emotion recognition and response generation given both the dialogue context and the situation description.

Let the dialogue context be
\begin{equation}
D = [U_1, U_2, \ldots, U_M]
\end{equation}
where $M$ denotes the number of dialogue turns. Each utterance is represented as
\begin{equation}
U_j = [w_j^1, w_j^2, \ldots, w_j^{m_j}]
\end{equation}
where $m_j$ is the number of tokens in the $j$-th utterance.

The situation description is defined as
\begin{equation}
S = [s_1, s_2, \ldots, s_L]
\end{equation}
where $L$ denotes the number of tokens.

To facilitate commonsense reasoning, we employ COMET~\cite{DBLP:conf/aaai/HwangBBDSBC21} to generate five types of commonsense inferences—xIntent, xNeed, xWant, xEffect, and xReact—for both the dialogue context and the situation description. 
Let $\mathcal{R}$ denote this set of relations. 
Further details and examples of COMET-based commonsense generation are provided in Appendix~\ref{sec:appendix_comet}.
We denote the corresponding commonsense sets as $K_D$ and $K_S$, respectively, and define
\begin{equation}
K = \{K_D, K_S\}
\end{equation}

Given $(D,S,K)$, the model jointly predicts the emotion label and generates an empathetic response. The response generation objective is
\begin{equation}
P(Y \mid D,S,K)=\prod_{t=1}^{N}P(y_t\mid y_{<t},D,S,K)
\end{equation}
where $Y=[y_1,y_2,\ldots,y_N]$ denotes the target response.

The emotion prediction objective is
\begin{equation}
\hat{e}=\arg\max_{e\in\mathcal{E}}P(e\mid D,S,K)
\end{equation}
where $\mathcal{E}$ denotes the 32 emotion categories in the Empathetic-Dialogues dataset. 
Both objectives are jointly optimized during training.

\subsection{Evaluation Metrics}\label{sec:evaluation_metrics}

Following previous commonsense-aware ERG studies~\cite{DBLP:conf/aaai/SabourZH22,DBLP:conf/aaai/LiLRRC22}, we evaluate our framework using four widely adopted automatic metrics.

\begin{itemize}
    \item \textbf{Perplexity (PPL)} measures how well the model predicts the reference responses. Lower values indicate better predictive performance.
    
    \item \textbf{Emotion Accuracy (Acc)} evaluates the model's ability to correctly predict the speaker's emotion label. Higher values indicate better emotion understanding.
    
    \item \textbf{Distinct-1/Distinct-2 (Dist-1/2)}~\cite{DBLP:conf/naacl/LiGBGD16} measure lexical diversity by computing the ratio of distinct unigrams and bigrams in the generated responses. Higher values indicate more diverse and less repetitive responses.
\end{itemize}

\section{Methodology}

\subsection{Framework Overview}
\begin{figure*}[t]
    \centering
    \includegraphics[width=0.8\textwidth]{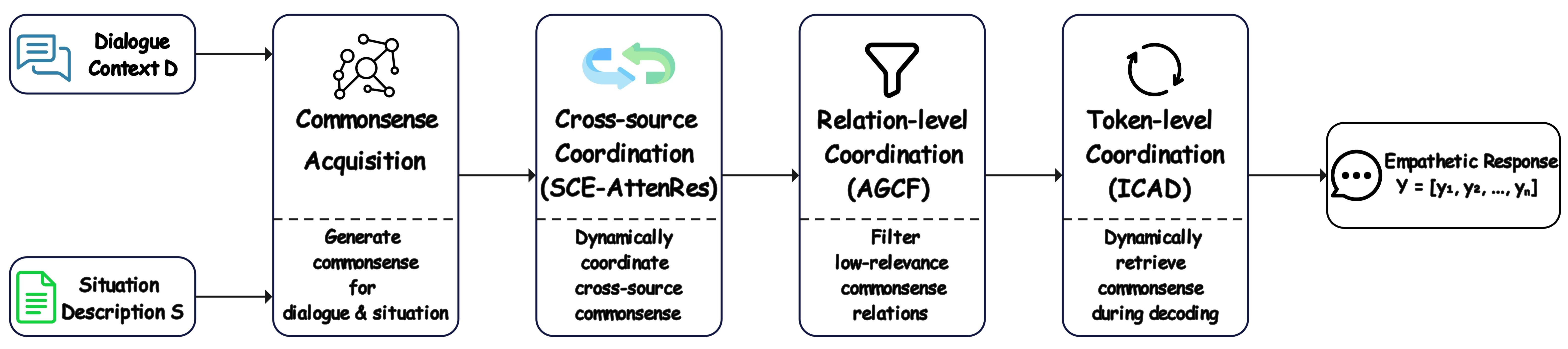}
    \caption{
    Overview of the proposed Dynamic Commonsense Coordination Framework. The framework coordinates commonsense utilization at three complementary levels: cross-source coordination through SCE-AttnRes, relation-level filtering through AGCF, and token-level retrieval through ICAD.
    Given the dialogue context $D$ and situation description $S$, the framework jointly supports emotion recognition and empathetic response generation.
    }
    \label{fig:framework}
\end{figure*}

Figure~\ref{fig:framework} presents the overall architecture of the proposed Dynamic Commonsense Coordination Framework. 
Given a dialogue context $D$ and its corresponding situation description $S$, the framework employs COMET to generate commonsense inferences for both sources, treating them as two independent commonsense streams.  
SCE-AttnRes first coordinates the two commonsense sources to refine dialogue understanding. The original commonsense representations are simultaneously organized into relation-level memories and filtered by AGCF. During response generation, ICAD dynamically retrieves the filtered memories at each decoding step.

The entire framework is jointly optimized for emotion recognition and empathetic response generation in an end-to-end manner.

Additional design motivations and implementation details are provided in Appendix~\ref{sec:appendix_method}.

\subsection{Commonsense Encoding and Cross-source Coordination (SCE-AttnRes)}\label{sec:sce_attnres}

\begin{figure*}[t]
    \centering
    \includegraphics[width=0.75\textwidth]{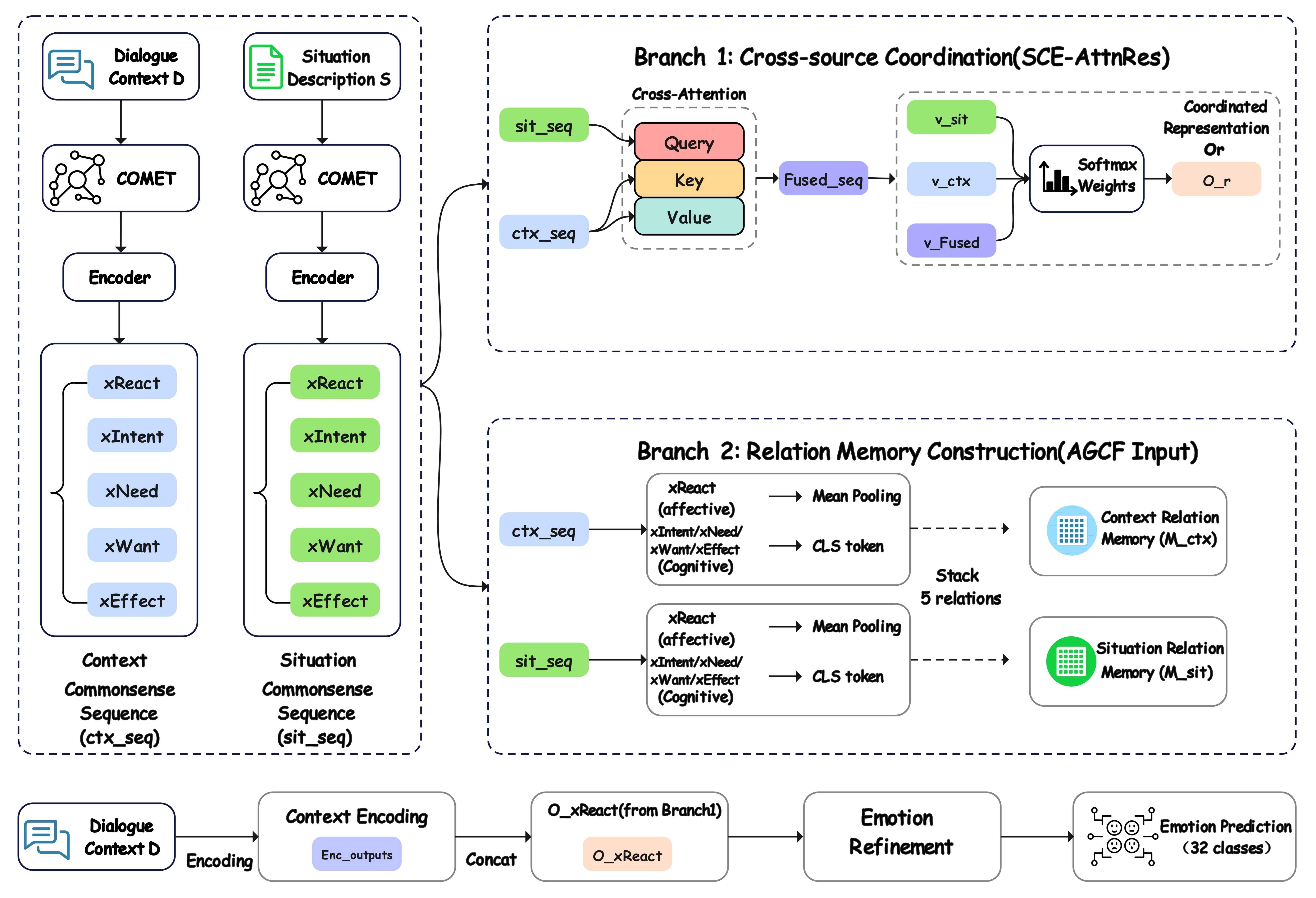}
    \caption{
    Overview of the SCE-AttnRes module for cross-source commonsense coordination. 
    Given dialogue $D$ and situation $S$, COMET generates five commonsense relations per source. 
    The two streams are independently encoded, fused via cross-attention (situation $\to$ dialogue), and adaptively balanced via attention-residual selection. 
    The coordinated representations are used for dialogue refinement, while the original encoded commonsense representations are preserved as relation-level memories for AGCF.}
    \label{fig:stage1}
\end{figure*}

Figure~\ref{fig:stage1} illustrates the SCE-AttnRes module, which consists of three steps: commonsense encoding, cross-source coordination, and attention-residual fusion.

\subsubsection{Commonsense Encoding}

COMET generates five relations for each source. The four cognitive relations are encoded using a shared cognitive encoder, while \textit{xReact} is encoded using a separate affective encoder. 
The same encoders are shared across the dialogue and situation streams. 
For each relation $r$, let
\begin{equation}
\mathbf{C}_r \in \mathbb{R}^{L_r \times d}
\end{equation}
and
\begin{equation}
\mathbf{S}_r \in \mathbb{R}^{L'_r \times d}
\end{equation}
denote the encoded dialogue and situation commonsense sequences, respectively. 

For the four cognitive relations, we use the \texttt{[CLS]} token:
\begin{equation}
\mathbf{h}^{ctx}_r = \mathbf{C}_r[0]; \quad \mathbf{h}^{sit}_r = \mathbf{S}_r[0]
\end{equation}
For \textit{xReact}, we apply mean pooling:
\begin{equation}
\mathbf{h}^{ctx}_{xReact} = \frac{1}{L_r}\sum_{i=1}^{L_r} \mathbf{C}_{xReact}[i]
\end{equation}
\begin{equation}
\mathbf{h}^{sit}_{xReact} = \frac{1}{L'_r}\sum_{i=1}^{L'_r} \mathbf{S}_{xReact}[i]
\end{equation}

\subsubsection{Cross-source Coordination}

To capture cross-source interactions, we apply cross-attention~\cite{DBLP:conf/nips/VaswaniSPUJGKP17} with situation-level commonsense as the query and dialogue-level as the key and value:
\begin{equation}
\mathbf{F}_r = \mathrm{CrossAttn}(\mathbf{S}_r, \mathbf{C}_r, \mathbf{C}_r)
\end{equation}

\subsubsection{Attention-Residual Fusion}

To preserve complementary information, we adaptively balance the dialogue source, situation source, and their fused representation. We compress each sequence via mean pooling:
\begin{equation}
\mathbf{v}_{ctx} = \mathrm{Mean}(\mathbf{C}_r)
\end{equation}
\begin{equation}
\mathbf{v}_{sit} = \mathrm{Mean}(\mathbf{S}_r)
\end{equation}
\begin{equation}
\mathbf{v}_{fused} = \mathrm{Mean}(\mathbf{F}_r)
\end{equation}

We compute an unnormalized score for each source:
\begin{equation}
s_i = \mathbf{w}^{\top}\mathrm{LayerNorm}(\mathbf{v}_i), \quad i \in \{ctx,sit,fused\}
\end{equation}
where $\mathbf{w}\in\mathbb{R}^{d}$ is a trainable parameter. 
The scores are then normalized by softmax:
\begin{equation}
\beta_i = \frac{\exp(s_i)}{\sum_j \exp(s_j)}
\end{equation}

The final coordinated representation is a weighted sum:
\begin{equation}
\mathbf{o}_r = \beta_{ctx}\mathbf{v}_{ctx} + \beta_{sit}\mathbf{v}_{sit} + \beta_{fused}\mathbf{v}_{fused}
\end{equation}

This procedure is independently applied to all five relations, yielding the coordinated representations
\begin{equation}
\{\mathbf{o}_{xIntent}, \mathbf{o}_{xNeed}, \mathbf{o}_{xWant}, \mathbf{o}_{xEffect}, \mathbf{o}_{xReact}\}.
\end{equation}

Following the refinement procedure of CEM~\cite{DBLP:conf/aaai/SabourZH22}, we concatenate $\mathbf{o}_{xReact}$ with each token of the original dialogue encoder output and pass the concatenated sequence through the affection-refinement encoder. 
Similarly, the four coordinated cognitive representations are concatenated and processed by the cognition-refinement encoder. 
The refined representations are then fused through a gating mechanism for decoding, while $\mathbf{o}_{xReact}$ is also used for emotion prediction.

In parallel, the original dialogue and situation commonsense representations are preserved as two relation-level memories for AGCF. Keeping the source-specific memories allows subsequent relation filtering to operate directly on the original commonsense sources rather than the already coordinated representations.

\subsection{Association-Guided Commonsense Filtering (AGCF)}\label{sec:agcf}

\begin{figure*}[t]
    \centering
    \includegraphics[width=0.75\textwidth]{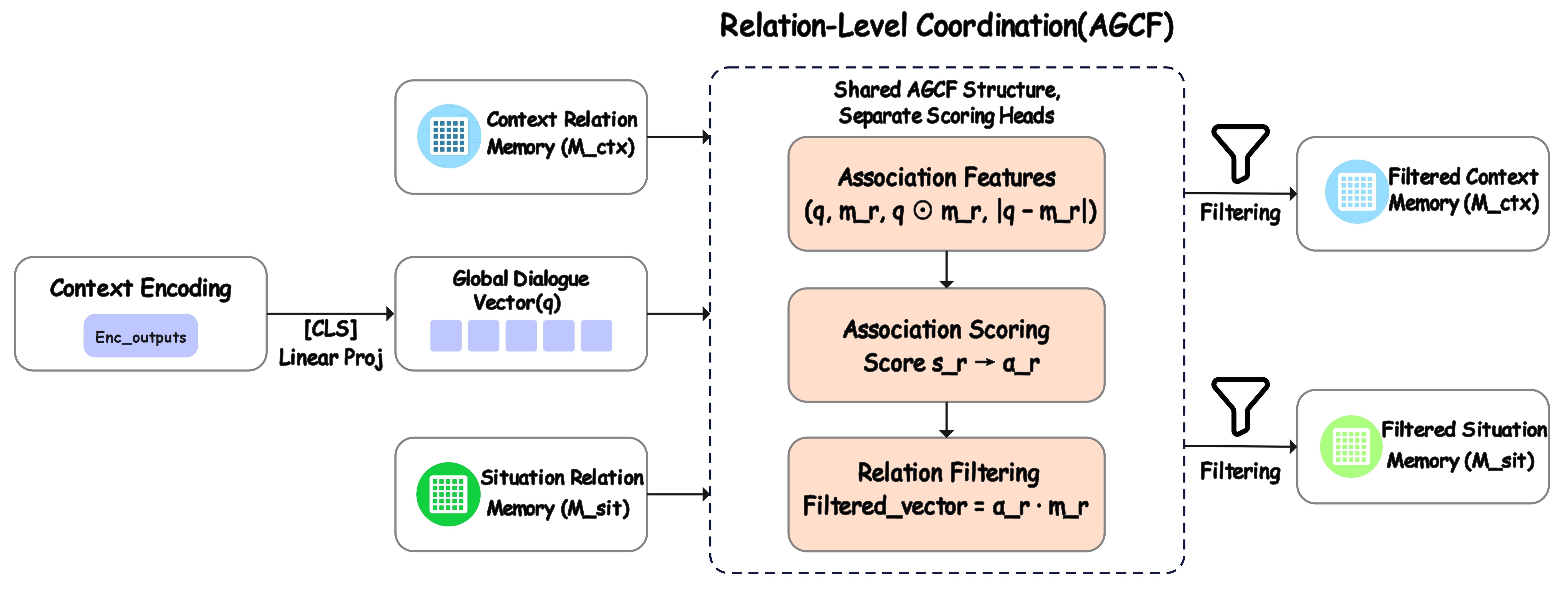}
    \caption{Overview of AGCF. Using the global dialogue representation and relation-level commonsense memories, AGCF computes association features per relation, estimates relevance via learnable scoring, and filters low-relevance commonsense independently for dialogue and situation memories.}
    \label{fig:stage2}
\end{figure*}

In parallel with cross-source coordination, we construct two relation-level commonsense memories from the independently encoded dialogue and situation commonsense representations: $\mathbf{M}^{ctx}, \mathbf{M}^{sit} \in \mathbb{R}^{5 \times d}$. 
Each memory contains one vector for each commonsense relation. 
AGCF then adaptively reweights these relations according to their association with the current dialogue, as shown in Figure~\ref{fig:stage2}.

We first obtain a global dialogue representation from the refined dialogue encoder output:
\begin{equation}
\mathbf{q} = \mathbf{W}_q \mathbf{h}_{CLS}
\end{equation}
where $\mathbf{h}_{CLS}$ is the \texttt{[CLS]} hidden state from the dialogue encoder output, and $\mathbf{W}_q\in\mathbb{R}^{d\times d}$ is a learnable projection matrix.

For each relation vector $\mathbf{m}_r$, we apply LayerNorm:
\begin{equation}
\bar{\mathbf{m}}_r = \mathrm{LayerNorm}(\mathbf{m}_r)
\end{equation}
and construct an association feature:
\begin{equation}
\mathbf{f}_r = [\mathbf{q}; \bar{\mathbf{m}}_r; \mathbf{q} \odot \bar{\mathbf{m}}_r; |\mathbf{q} - \bar{\mathbf{m}}_r|]
\end{equation}

Separate linear scoring heads are used for the dialogue and situation memories. For source $z\in\{ctx,sit\}$, the association score is computed as:
\begin{equation}
s_r^{z} = \mathbf{W}_s^{z}\mathbf{f}_r^{z}, \qquad z\in\{ctx,sit\}
\end{equation}

The relevance weights are obtained via softmax with rescaling:
\begin{equation}
\alpha_r^{z} = \frac{\exp(s_r^{z})}{\sum_j \exp(s_j^{z})} \cdot |R|, \qquad z\in\{ctx,sit\}
\end{equation}
where $|R|=5$ is the number of commonsense relations. The rescaling factor prevents the softmax weights from uniformly shrinking all relation contributions.

Each commonsense relation is filtered according to its relevance weight:
\begin{equation}
\tilde{\mathbf{m}}_r^{z} = \alpha_r^{z} \bar{\mathbf{m}}_r^{z}, \qquad z\in\{ctx,sit\}
\end{equation}

This procedure is independently applied to both memories, producing the filtered memories $\tilde{\mathbf{M}}^{ctx}$ and $\tilde{\mathbf{M}}^{sit}$ for decoding.

\subsection{Iterative Commonsense-Aware Decoding (ICAD)}\label{sec:icad}

\begin{figure*}[t]
    \centering
    \includegraphics[width=0.75\textwidth]{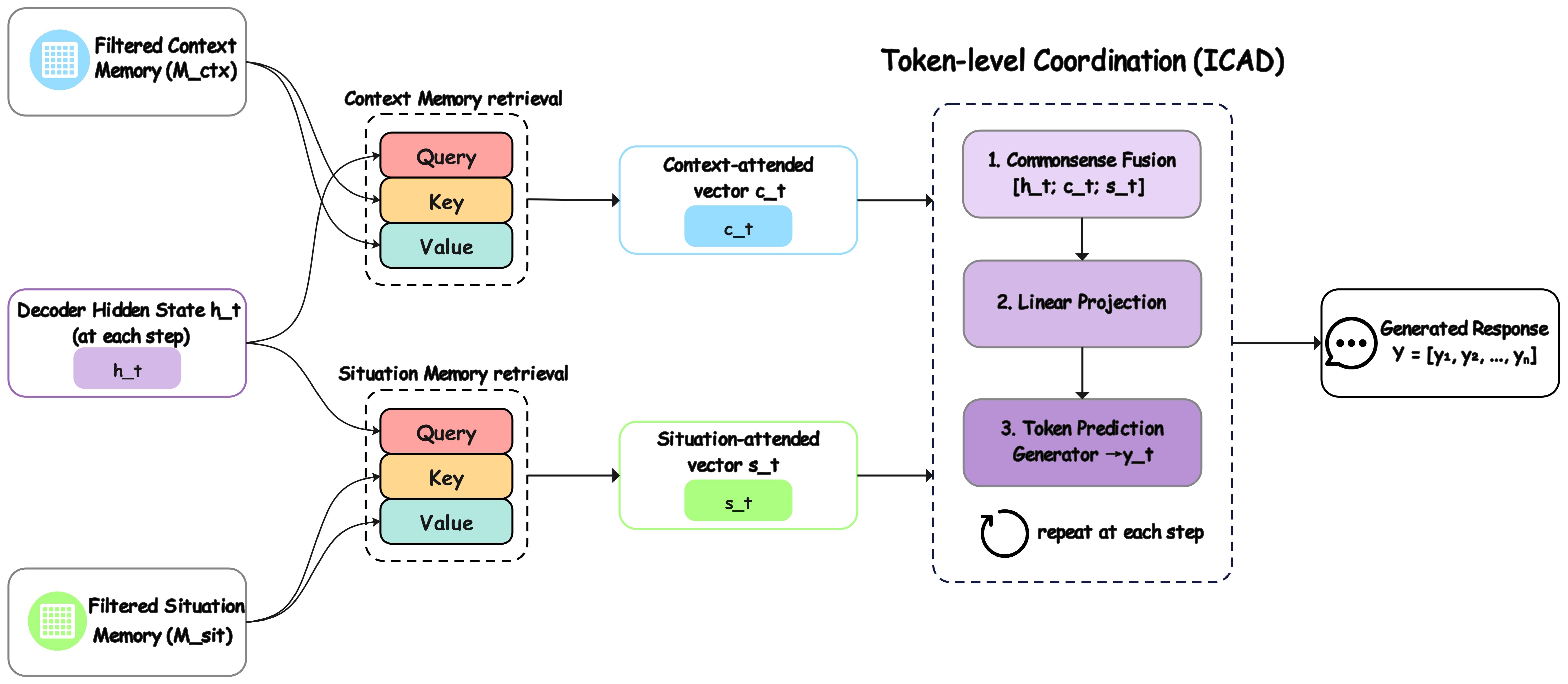}
    \caption{Overview of ICAD. At each decoding step, the decoder hidden state retrieves relevant information from filtered dialogue and situation commonsense memories via cross-attention. The retrieved representations are fused with the decoder state to predict the next token, enabling dynamic commonsense utilization during decoding.}
    \label{fig:stage3}
\end{figure*}

ICAD dynamically retrieves commonsense at every decoding step from the filtered memories $\tilde{\mathbf{M}}^{ctx}$ and $\tilde{\mathbf{M}}^{sit}$, as shown in Figure~\ref{fig:stage3}.

We place ICAD after the decoder's self-attention sub-layer. 
Let $\mathbf{h}_t^{dec}\in\mathbb{R}^{d}$ denote the decoder hidden state at step $t$ after self-attention, which serves as the query for commonsense retrieval. 
We retrieve dialogue and situation commonsense via cross-attention:
\begin{equation}
\mathbf{c}_t = \mathrm{CrossAttn}(\mathbf{h}_t^{dec}, \tilde{\mathbf{M}}^{ctx}, \tilde{\mathbf{M}}^{ctx})
\end{equation}
\begin{equation}
\mathbf{s}_t = \mathrm{CrossAttn}(\mathbf{h}_t^{dec}, \tilde{\mathbf{M}}^{sit}, \tilde{\mathbf{M}}^{sit})
\end{equation}

These are fused with the decoder state:
\begin{equation}
\tilde{\mathbf{h}}_t^{dec} = \mathbf{W}_f[\mathbf{h}_t^{dec}; \mathbf{c}_t; \mathbf{s}_t]
\end{equation}
where $\mathbf{W}_f\in\mathbb{R}^{d\times3d}$ is a learnable projection matrix.

The fused representation $\tilde{\mathbf{h}}_t^{dec}$ is used to predict the next token. Unlike static commonsense injection, ICAD performs retrieval at every step. Because the decoder query changes across decoding steps, ICAD can assign different attention distributions to the fixed relation memories as the response unfolds.

\subsection{Training Objectives}\label{sec:training}

We adopt the same multi-task training setup as CEM~\cite{DBLP:conf/aaai/SabourZH22}, jointly optimizing for emotion classification, generation, and diversity. 
The refined emotion-aware representation $\mathbf{h}_{emo}$, obtained by passing $\mathbf{o}_{xReact}$ through the affection-refinement encoder, is used for emotion classification loss:

\begin{equation}
\mathcal{L}_{emo}
=
-\log P(e^* \mid \mathbf{h}_{emo})
\end{equation}

The response generation loss is the standard negative log-likelihood:

\begin{equation}
\mathcal{L}_{nll}
=
-\sum_{t=1}^{N}
\log P(y_t \mid y_{<t},D,S,K)
\end{equation}

To encourage diverse generation, we adopt the Frequency-Aware Cross-Entropy (FACE) loss~\cite{DBLP:conf/www/JiangRMR19}:

\begin{equation}
\mathcal{L}_{div}
=
-\sum_{t=1}^{N}
w_{y_t}
\log P(y_t \mid y_{<t},D,S,K)
\end{equation}

where $w_{y_t}$ is the frequency-aware weight computed from prediction frequencies accumulated during training.

The overall objective is

\begin{equation}
\mathcal{L}
=
\mathcal{L}_{nll}
+
\mathcal{L}_{emo}
+
\gamma\mathcal{L}_{div}
\end{equation}

where $\gamma$ controls the contribution of the diversity objective (we use $\gamma=2.0$ in our experiments, while the original CEM uses $1.5$). All components are jointly optimized in an end-to-end manner.

\section{Experiments}

\subsection{Experimental Settings}

We conduct experiments on the Empathetic-Dialogues dataset~\cite{DBLP:conf/acl/RashkinSLB19}. Our framework is implemented based on the official CEM implementation~\cite{DBLP:conf/aaai/SabourZH22} to ensure fair comparison. Detailed dataset statistics, hyperparameter configurations, and training procedures are provided in Appendix~\ref{sec:appendix_experimental_settings}.

For brevity, we denote our proposed Dynamic Commonsense Coordination framework as \textbf{DCC} in the following experiments.

Following the evaluation protocol described in Section~\ref{sec:evaluation_metrics}, we report Perplexity (PPL), Distinct-1/2, and emotion classification accuracy (Acc) for all experiments.

\subsection{Comparison with Baselines}

We compare DCC with emotion-aware models
MoEL~\cite{DBLP:conf/emnlp/LinMSXF19},
MIME~\cite{DBLP:conf/emnlp/MajumderHPLGGMP20}, and
EmpDG~\cite{DBLP:conf/coling/LiCRRTC20};
commonsense-aware models
KEMP~\cite{DBLP:conf/aaai/LiLRRC22},
CEM~\cite{DBLP:conf/aaai/SabourZH22},
SEEK~\cite{DBLP:conf/emnlp/WangLLMYWZ22}, and
CASE~\cite{DBLP:conf/acl/ZhouZW0H23}; and recent models
ESCM~\cite{DBLP:conf/emnlp/YangRYZCCWSJL23},
E-CORE~\cite{DBLP:conf/emnlp/FuZWM23},
SDAM~\cite{DBLP:journals/ipm/YangRWSZL24}, and
IAMM~\cite{DBLP:conf/acl/YangRYSCZL24}.

Table~\ref{tab:baseline_comparison} presents the main results on the
Empathetic-Dialogues~\cite{DBLP:conf/acl/RashkinSLB19} test set.

\begin{table}[t]
\centering
\caption{Comparison with baseline models on the Empathetic-Dialogues
test set. The best and second-best results are shown in bold and
underlined, respectively. Results marked with $^\dagger$ are taken
from the corresponding original papers.}
\label{tab:baseline_comparison}
\setlength{\tabcolsep}{4.5pt}
\renewcommand{\arraystretch}{1.10}
\normalsize
\begin{tabular}{lcccc}
\toprule
\textbf{Model}
& \textbf{PPL} $\downarrow$
& \textbf{Dist-1} $\uparrow$
& \textbf{Dist-2} $\uparrow$
& \textbf{Acc} $\uparrow$ \\
\midrule
MoEL$^\dagger$   & 38.04 & 0.44 & 2.10 & 32.00 \\
MIME$^\dagger$   & 37.09 & 0.47 & 1.91 & 34.24 \\
EmpDG$^\dagger$  & 37.29 & 0.46 & 2.02 & 34.31 \\
KEMP$^\dagger$   & 36.89 & 0.55 & 2.29 & 39.31 \\
CEM$^\dagger$    & 36.11 & 0.66 & 2.99 & 39.11 \\
SEEK$^\dagger$   & 37.09 & 0.73 & 3.23 & 41.85 \\
CASE$^\dagger$   & 35.37 & 0.74 & 4.01 & 40.20 \\
ESCM$^\dagger$   & \underline{34.82} & 1.19
                  & 4.11 & 41.19 \\
E-CORE$^\dagger$ & \textbf{33.03} & 0.72
                  & 3.49 & 42.59 \\
SDAM$^\dagger$   & 35.07 & \underline{1.60}
                  & \underline{5.24} & \underline{52.45} \\
IAMM$^\dagger$   & 35.66 & \textbf{2.09}
                  & \textbf{7.03} & \textbf{55.92} \\
\midrule
\textbf{DCC}     & 36.13 & 1.03 & 4.93 & 46.09 \\
\bottomrule
\end{tabular}
\end{table}

IAMM and SDAM achieve the strongest overall results, reflecting the benefit of their fine-grained associative modeling. The most direct comparison for DCC is with CEM, since DCC is built directly on its official codebase\footnote{\url{https://github.com/Sahandfer/CEM}}. 
DCC improves Acc from
39.11\% to 46.09\%, Dist-1 from 0.66 to 1.03, and Dist-2 from 2.99
to 4.93, while maintaining nearly identical PPL (36.13 versus 36.11).
The gains over CEM suggest that coordinating commonsense at the source, relation, and decoding levels improves both emotion prediction and response diversity. 
Detailed comparisons with IAMM and SDAM are provided in
Appendix~\ref{sec:comparison_recent_models}.

\subsection{Ablation Study}

Table~\ref{tab:ablation} evaluates the three DCC components.
Implementation details and extended analysis of the ablation variants
are provided in Appendix~\ref{sec:ablation_details}.

\begin{table}[t]
\centering
\caption{Ablation results on the Empathetic-Dialogues test set.
The best result in each column is shown in bold.}
\label{tab:ablation}
\setlength{\tabcolsep}{5pt}
\renewcommand{\arraystretch}{1.15}
\small
\begin{tabular}{lcccc}
\toprule
\textbf{Model}
& \textbf{PPL} $\downarrow$
& \textbf{Dist-1} $\uparrow$
& \textbf{Dist-2} $\uparrow$
& \textbf{Acc} $\uparrow$ \\
\midrule
\textbf{DCC}
                 & \textbf{36.13}
                 & 1.03
                 & 4.93
                 & 46.09 \\
\midrule
w/o SCE-AttnRes  & 36.89 & 0.77 & 3.88 & 36.59 \\
w/o AGCF          & 37.72 & 0.73 & 3.66 & \textbf{46.89} \\
w/o ICAD          & 37.48 & \textbf{1.05}
                  & \textbf{5.59} & 44.99 \\
\bottomrule
\end{tabular}
\end{table}

Removing SCE-AttnRes causes the largest Acc drop (from 46.09\% to 36.59\%) and also degrades PPL and diversity, confirming the importance of cross-source coordination. 
Removing AGCF, by contrast, yields the worst PPL and diversity while slightly improving Acc to 46.89\%, suggesting that AGCF mainly benefits generation. The Acc increase likely reflects retained affective signals that aid classification, even as they impair generation.
Replacing ICAD with static aggregation lowers PPL and Acc but raises Dist-1/2, revealing a trade-off between dynamic relevance and surface-level diversity. 

Overall, SCE-AttnRes primarily supports emotion understanding, AGCF filters relation-level knowledge for generation, and ICAD dynamically controls its use during decoding.

\subsection{Additional Analyses}

AGCF and ICAD visualizations are provided in Appendix~\ref{sec:agcf_visualization} and Appendix~\ref{sec:icad_visualization}, respectively. 
A blind LLM-based comparison (Appendix~\ref{sec:llm_evaluation}) shows that DCC improves relevance, coherence, and informativeness over CEM, while empathy and overall gains are not significant. 
A human preference study is reported in Appendix~\ref{sec:human_evaluation}. 
Efficiency measurements (Appendix~\ref{sec:efficiency_analysis}) show that DCC increases latency by 32.55\% while retaining 9.00 samples/sec on an RTX 4090.

\section{Conclusion}

We proposed DCC, a Dynamic Commonsense Coordination framework for empathetic response generation. 
DCC coordinates commonsense across dialogue understanding and response generation through three complementary modules: SCE-AttnRes for cross-source integration, AGCF for relation-level filtering, and ICAD for token-level retrieval. 
Experiments on Empathetic-Dialogues show that DCC consistently improves emotion classification accuracy and response diversity over the CEM baseline while maintaining comparable perplexity. 
Ablation and mechanism analyses confirm the distinct roles of the three modules, and an LLM-based evaluation further indicates improvements in relevance, coherence, and informativeness. A small-scale human preference study also shows consistent preference toward DCC, particularly for empathy.

Future work will explore integrating DCC with large language models and developing more fine-grained commonsense selection mechanisms that can identify and update relevant knowledge beyond predefined relation sets.

\section*{Limitations}

Our experiments are limited to the English Empathetic-Dialogues dataset, leaving generalization to other languages, domains, and longer conversations unexamined. The crowd-sourced dialogues are relatively short and may not capture the full complexity of real-world empathy.

DCC also inherits limitations from COMET, which supplies commonsense only across five predefined relations. COMET outputs can be noisy or biased, and knowledge outside this fixed set is inaccessible. The strong suppression of \textit{xReact} in our analysis further suggests that the affective-cognitive commonsense interaction needs more investigation.

We also acknowledge that DCC does not outperform recent association-based models like IAMM and SDAM on automatic metrics. Our contribution is to demonstrate the effects of coordinating commonsense across sources, relations, and decoding steps within CEM, rather than to claim a new state of the art.

Finally, the ablation results are single-run and lack statistical testing. Small differences, such as the Acc change after removing AGCF, may reflect training variation. Automatic and LLM-based evaluations also cannot fully assess genuine empathy. 
Although we conduct a small-scale human preference study, a larger evaluation with more annotators is needed for more reliable assessment.

\section*{Ethical Considerations}

We use the publicly available Empathetic-Dialogues dataset and did not collect any new user data. 
However, both the dataset and COMET may carry cultural, demographic, or social biases from the crowd-sourced data and underlying training corpora. 
DCC could therefore generate stereotypical, inappropriate, or emotionally tone-deaf responses.

It would be misleading to treat empathetic dialogue systems as if they understood emotions. They are not a substitute for qualified human support in mental-health or crisis situations. 
Misreading a user's emotional state or providing overconfident advice could cause harm in high-stakes situations. 
Practical deployment would require safety filtering, uncertainty-aware response strategies, and mechanisms for escalation to human professionals.

Our LLM-based evaluation is only supplementary. 
LLM judges may be influenced by response position, wording, style, or cultural assumptions. 
We tried to mitigate some of these effects through anonymization and randomized ordering, but these judgments should not be considered equivalent to human evaluation.

The human evaluation was conducted with informed consent, anonymized responses, and no collection of personally identifying information.

\bibliography{custom}

@article{davis1983measuring,
  title={Measuring individual differences in empathy: Evidence for a multidimensional approach.},
  author={Davis, Mark H},
  journal={Journal of personality and social psychology},
  volume={44},
  number={1},
  pages={113},
  year={1983},
  publisher={American Psychological Association}
}

@article{smith2006cognitive,
  title={Cognitive empathy and emotional empathy in human behavior and evolution},
  author={Smith, Adam},
  journal={The Psychological Record},
  volume={56},
  number={1},
  pages={3--21},
  year={2006},
  publisher={Springer}
}

@article{singer2009social,
  title={The social neuroscience of empathy},
  author={Singer, Tania and Lamm, Claus},
  journal={Annals of the new York Academy of Sciences},
  volume={1156},
  number={1},
  pages={81--96},
  year={2009},
  publisher={Wiley Online Library}
}

@inproceedings{DBLP:conf/acl/RashkinSLB19,
  author       = {Hannah Rashkin and
                  Eric Michael Smith and
                  Margaret Li and
                  Y{-}Lan Boureau},
  editor       = {Anna Korhonen and
                  David R. Traum and
                  Llu{\'{\i}}s M{\`{a}}rquez},
  title        = {Towards Empathetic Open-domain Conversation Models: {A} New Benchmark
                  and Dataset},
  booktitle    = {Proceedings of the 57th Conference of the Association for Computational
                  Linguistics, {ACL} 2019, Florence, Italy, July 28- August 2, 2019,
                  Volume 1: Long Papers},
  pages        = {5370--5381},
  publisher    = {Association for Computational Linguistics},
  year         = {2019},
  url          = {https://doi.org/10.18653/v1/p19-1534},
  doi          = {10.18653/V1/P19-1534},
  bibsource    = {dblp computer science bibliography, https://dblp.org}
}

@inproceedings{DBLP:conf/emnlp/LinMSXF19,
  author       = {Zhaojiang Lin and
                  Andrea Madotto and
                  Jamin Shin and
                  Peng Xu and
                  Pascale Fung},
  editor       = {Kentaro Inui and
                  Jing Jiang and
                  Vincent Ng and
                  Xiaojun Wan},
  title        = {MoEL: Mixture of Empathetic Listeners},
  booktitle    = {Proceedings of the 2019 Conference on Empirical Methods in Natural
                  Language Processing and the 9th International Joint Conference on
                  Natural Language Processing, {EMNLP-IJCNLP} 2019, Hong Kong, China,
                  November 3-7, 2019},
  pages        = {121--132},
  publisher    = {Association for Computational Linguistics},
  year         = {2019},
  url          = {https://doi.org/10.18653/v1/D19-1012},
  doi          = {10.18653/V1/D19-1012},
  bibsource    = {dblp computer science bibliography, https://dblp.org}
}

@inproceedings{DBLP:conf/emnlp/MajumderHPLGGMP20,
  author       = {Navonil Majumder and
                  Pengfei Hong and
                  Shanshan Peng and
                  Jiankun Lu and
                  Deepanway Ghosal and
                  Alexander F. Gelbukh and
                  Rada Mihalcea and
                  Soujanya Poria},
  editor       = {Bonnie Webber and
                  Trevor Cohn and
                  Yulan He and
                  Yang Liu},
  title        = {{MIME:} MIMicking Emotions for Empathetic Response Generation},
  booktitle    = {Proceedings of the 2020 Conference on Empirical Methods in Natural
                  Language Processing, {EMNLP} 2020, Online, November 16-20, 2020},
  pages        = {8968--8979},
  publisher    = {Association for Computational Linguistics},
  year         = {2020},
  url          = {https://doi.org/10.18653/v1/2020.emnlp-main.721},
  doi          = {10.18653/V1/2020.EMNLP-MAIN.721},
  bibsource    = {dblp computer science bibliography, https://dblp.org}
}

@inproceedings{DBLP:conf/coling/LiCRRTC20,
  author       = {Qintong Li and
                  Hongshen Chen and
                  Zhaochun Ren and
                  Pengjie Ren and
                  Zhaopeng Tu and
                  Zhumin Chen},
  editor       = {Donia Scott and
                  N{\'{u}}ria Bel and
                  Chengqing Zong},
  title        = {EmpDG: Multi-resolution Interactive Empathetic Dialogue Generation},
  booktitle    = {Proceedings of the 28th International Conference on Computational
                  Linguistics, {COLING} 2020, Barcelona, Spain (Online), December 8-13,
                  2020},
  pages        = {4454--4466},
  publisher    = {International Committee on Computational Linguistics},
  year         = {2020},
  url          = {https://doi.org/10.18653/v1/2020.coling-main.394},
  doi          = {10.18653/V1/2020.COLING-MAIN.394},
  bibsource    = {dblp computer science bibliography, https://dblp.org}
}

@inproceedings{DBLP:conf/aaai/SabourZH22,
  author       = {Sahand Sabour and
                  Chujie Zheng and
                  Minlie Huang},
  title        = {{CEM:} Commonsense-Aware Empathetic Response Generation},
  booktitle    = {Thirty-Sixth {AAAI} Conference on Artificial Intelligence, {AAAI}
                  2022, Thirty-Fourth Conference on Innovative Applications of Artificial
                  Intelligence, {IAAI} 2022, The Twelveth Symposium on Educational Advances
                  in Artificial Intelligence, {EAAI} 2022 Virtual Event, February 22
                  - March 1, 2022},
  pages        = {11229--11237},
  publisher    = {{AAAI} Press},
  year         = {2022},
  url          = {https://doi.org/10.1609/aaai.v36i10.21373},
  doi          = {10.1609/AAAI.V36I10.21373},
  bibsource    = {dblp computer science bibliography, https://dblp.org}
}

@inproceedings{DBLP:conf/aaai/LiLRRC22,
  author       = {Qintong Li and
                  Piji Li and
                  Zhaochun Ren and
                  Pengjie Ren and
                  Zhumin Chen},
  title        = {Knowledge Bridging for Empathetic Dialogue Generation},
  booktitle    = {Thirty-Sixth {AAAI} Conference on Artificial Intelligence, {AAAI}
                  2022, Thirty-Fourth Conference on Innovative Applications of Artificial
                  Intelligence, {IAAI} 2022, The Twelveth Symposium on Educational Advances
                  in Artificial Intelligence, {EAAI} 2022 Virtual Event, February 22
                  - March 1, 2022},
  pages        = {10993--11001},
  publisher    = {{AAAI} Press},
  year         = {2022},
  url          = {https://doi.org/10.1609/aaai.v36i10.21347},
  doi          = {10.1609/AAAI.V36I10.21347},
  bibsource    = {dblp computer science bibliography, https://dblp.org}
}

@inproceedings{DBLP:conf/emnlp/WangLLMYWZ22,
  author       = {Lanrui Wang and
                  Jiangnan Li and
                  Zheng Lin and
                  Fandong Meng and
                  Chenxu Yang and
                  Weiping Wang and
                  Jie Zhou},
  editor       = {Yoav Goldberg and
                  Zornitsa Kozareva and
                  Yue Zhang},
  title        = {Empathetic Dialogue Generation via Sensitive Emotion Recognition and
                  Sensible Knowledge Selection},
  booktitle    = {Findings of the Association for Computational Linguistics: {EMNLP}
                  2022, Abu Dhabi, United Arab Emirates, December 7-11, 2022},
  series       = {Findings of {ACL}},
  pages        = {4634--4645},
  publisher    = {Association for Computational Linguistics},
  year         = {2022},
  url          = {https://doi.org/10.18653/v1/2022.findings-emnlp.340},
  doi          = {10.18653/V1/2022.FINDINGS-EMNLP.340},
  bibsource    = {dblp computer science bibliography, https://dblp.org}
}

@inproceedings{DBLP:conf/acl/ZhouZW0H23,
  author       = {Jinfeng Zhou and
                  Chujie Zheng and
                  Bo Wang and
                  Zheng Zhang and
                  Minlie Huang},
  editor       = {Anna Rogers and
                  Jordan L. Boyd{-}Graber and
                  Naoaki Okazaki},
  title        = {{CASE:} Aligning Coarse-to-Fine Cognition and Affection for Empathetic
                  Response Generation},
  booktitle    = {Proceedings of the 61st Annual Meeting of the Association for Computational
                  Linguistics (Volume 1: Long Papers), {ACL} 2023, Toronto, Canada,
                  July 9-14, 2023},
  pages        = {8223--8237},
  publisher    = {Association for Computational Linguistics},
  year         = {2023},
  url          = {https://doi.org/10.18653/v1/2023.acl-long.457},
  doi          = {10.18653/V1/2023.ACL-LONG.457},
  bibsource    = {dblp computer science bibliography, https://dblp.org}
}

@inproceedings{DBLP:conf/emnlp/FuZWM23,
  author       = {Fengyi Fu and
                  Lei Zhang and
                  Quan Wang and
                  Zhendong Mao},
  editor       = {Houda Bouamor and
                  Juan Pino and
                  Kalika Bali},
  title        = {{E-CORE:} Emotion Correlation Enhanced Empathetic Dialogue Generation},
  booktitle    = {Proceedings of the 2023 Conference on Empirical Methods in Natural
                  Language Processing, {EMNLP} 2023, Singapore, December 6-10, 2023},
  pages        = {10568--10586},
  publisher    = {Association for Computational Linguistics},
  year         = {2023},
  url          = {https://doi.org/10.18653/v1/2023.emnlp-main.653},
  doi          = {10.18653/V1/2023.EMNLP-MAIN.653},
  bibsource    = {dblp computer science bibliography, https://dblp.org}
}

@inproceedings{DBLP:conf/acl/CaiSXSWGZX23,
  author       = {Hua Cai and
                  Xuli Shen and
                  Qing Xu and
                  Weilin Shen and
                  Xiaomei Wang and
                  Weifeng Ge and
                  Xiaoqing Zheng and
                  Xiangyang Xue},
  editor       = {Anna Rogers and
                  Jordan L. Boyd{-}Graber and
                  Naoaki Okazaki},
  title        = {Improving Empathetic Dialogue Generation by Dynamically Infusing Commonsense
                  Knowledge},
  booktitle    = {Findings of the Association for Computational Linguistics: {ACL} 2023,
                  Toronto, Canada, July 9-14, 2023},
  series       = {Findings of {ACL}},
  pages        = {7858--7873},
  publisher    = {Association for Computational Linguistics},
  year         = {2023},
  url          = {https://doi.org/10.18653/v1/2023.findings-acl.498},
  doi          = {10.18653/V1/2023.FINDINGS-ACL.498},
  bibsource    = {dblp computer science bibliography, https://dblp.org}
}

@inproceedings{DBLP:conf/emnlp/YangRYZCCWSJL23,
  author       = {Zhou Yang and
                  Zhaochun Ren and
                  Yufeng Wang and
                  Xiaofei Zhu and
                  Zhihao Chen and
                  Tiecheng Cai and
                  Yunbing Wu and
                  Yisong Su and
                  Sibo Ju and
                  Xiangwen Liao},
  editor       = {Houda Bouamor and
                  Juan Pino and
                  Kalika Bali},
  title        = {Exploiting Emotion-Semantic Correlations for Empathetic Response Generation},
  booktitle    = {Findings of the Association for Computational Linguistics: {EMNLP}
                  2023, Singapore, December 6-10, 2023},
  series       = {Findings of {ACL}},
  pages        = {4826--4837},
  publisher    = {Association for Computational Linguistics},
  year         = {2023},
  url          = {https://doi.org/10.18653/v1/2023.findings-emnlp.320},
  doi          = {10.18653/V1/2023.FINDINGS-EMNLP.320},
  bibsource    = {dblp computer science bibliography, https://dblp.org}
}

@article{DBLP:journals/ipm/YangRWSZL24,
  author       = {Zhou Yang and
                  Zhaochun Ren and
                  Yufeng Wang and
                  Haizhou Sun and
                  Xiaofei Zhu and
                  Xiangwen Liao},
  title        = {Situation-aware empathetic response generation},
  journal      = {Inf. Process. Manag.},
  volume       = {61},
  number       = {6},
  pages        = {103824},
  year         = {2024},
  url          = {https://doi.org/10.1016/j.ipm.2024.103824},
  doi          = {10.1016/J.IPM.2024.103824},
  bibsource    = {dblp computer science bibliography, https://dblp.org}
}

@inproceedings{DBLP:conf/acl/YangRYSCZL24,
  author       = {Zhou Yang and
                  Zhaochun Ren and
                  Yufeng Wang and
                  Haizhou Sun and
                  Chao Chen and
                  Xiaofei Zhu and
                  Xiangwen Liao},
  editor       = {Lun{-}Wei Ku and
                  Andre Martins and
                  Vivek Srikumar},
  title        = {An Iterative Associative Memory Model for Empathetic Response Generation},
  booktitle    = {Proceedings of the 62nd Annual Meeting of the Association for Computational
                  Linguistics (Volume 1: Long Papers), {ACL} 2024, Bangkok, Thailand,
                  August 11-16, 2024},
  pages        = {3081--3092},
  publisher    = {Association for Computational Linguistics},
  year         = {2024},
  url          = {https://doi.org/10.18653/v1/2024.acl-long.170},
  doi          = {10.18653/V1/2024.ACL-LONG.170},
  bibsource    = {dblp computer science bibliography, https://dblp.org}
}

@article{DBLP:journals/corr/abs-2408-11599,
  author       = {Xinhao Chen and
                  Chong Yang and
                  Man Lan and
                  Li Cai and
                  Yang Chen and
                  Tu Hu and
                  Xinlin Zhuang and
                  Aimin Zhou},
  title        = {Cause-Aware Empathetic Response Generation via Chain-of-Thought Fine-Tuning},
  journal      = {CoRR},
  volume       = {abs/2408.11599},
  year         = {2024},
  url          = {https://doi.org/10.48550/arXiv.2408.11599},
  doi          = {10.48550/ARXIV.2408.11599},
  eprinttype   = {arXiv},
  eprint       = {2408.11599},
  bibsource    = {dblp computer science bibliography, https://dblp.org}
}

@inproceedings{DBLP:conf/cikm/Ren0YSCZL25,
  author       = {Zhaochun Ren and
                  Zhou Yang and
                  Chenglong Ye and
                  Haizhou Sun and
                  Chao Chen and
                  Xiaofei Zhu and
                  Xiangwen Liao},
  editor       = {Meeyoung Cha and
                  Chanyoung Park and
                  Noseong Park and
                  Carl Yang and
                  Senjuti Basu Roy and
                  Jessie Li and
                  Jaap Kamps and
                  Kijung Shin and
                  Bryan Hooi and
                  Lifang He},
  title        = {Fine-Grained Emotion Recognition via In-Context Learning},
  booktitle    = {Proceedings of the 34th {ACM} International Conference on Information
                  and Knowledge Management, {CIKM} 2025, Seoul, Republic of Korea, November
                  10-14, 2025},
  pages        = {2503--2513},
  publisher    = {{ACM}},
  year         = {2025},
  url          = {https://doi.org/10.1145/3746252.3761319},
  doi          = {10.1145/3746252.3761319},
  bibsource    = {dblp computer science bibliography, https://dblp.org}
}

@inproceedings{DBLP:conf/acl/YuanDCYN25,
  author       = {Jiahao Yuan and
                  Zixiang Di and
                  Zhiqing Cui and
                  Guisong Yang and
                  Usman Naseem},
  editor       = {Wanxiang Che and
                  Joyce Nabende and
                  Ekaterina Shutova and
                  Mohammad Taher Pilehvar},
  title        = {ReflectDiffu: Reflect between Emotion-intent Contagion and Mimicry
                  for Empathetic Response Generation via a RL-Diffusion Framework},
  booktitle    = {Proceedings of the 63rd Annual Meeting of the Association for Computational
                  Linguistics (Volume 1: Long Papers), {ACL} 2025, Vienna, Austria,
                  July 27 - August 1, 2025},
  pages        = {25435--25449},
  publisher    = {Association for Computational Linguistics},
  year         = {2025},
  url          = {https://doi.org/10.18653/v1/2025.acl-long.1235},
  doi          = {10.18653/V1/2025.ACL-LONG.1235},
  bibsource    = {dblp computer science bibliography, https://dblp.org}
}

@article{DBLP:journals/corr/abs-2604-07100,
  author       = {Hongru Ji and
                  Yuyin Fan and
                  Meng Zhao and
                  Xianghua Li and
                  Lianwei Wu and
                  Chao Gao},
  title        = {{STRIDE-ED:} {A} Strategy-Grounded Stepwise Reasoning Framework for
                  Empathetic Dialogue Systems},
  journal      = {CoRR},
  volume       = {abs/2604.07100},
  year         = {2026},
  url          = {https://doi.org/10.48550/arXiv.2604.07100},
  doi          = {10.48550/ARXIV.2604.07100},
  eprinttype   = {arXiv},
  eprint       = {2604.07100},
  bibsource    = {dblp computer science bibliography, https://dblp.org}
}

@inproceedings{DBLP:conf/acl/BosselutRSMCC19,
  author       = {Antoine Bosselut and
                  Hannah Rashkin and
                  Maarten Sap and
                  Chaitanya Malaviya and
                  Asli Celikyilmaz and
                  Yejin Choi},
  editor       = {Anna Korhonen and
                  David R. Traum and
                  Llu{\'{\i}}s M{\`{a}}rquez},
  title        = {{COMET:} Commonsense Transformers for Automatic Knowledge Graph Construction},
  booktitle    = {Proceedings of the 57th Conference of the Association for Computational
                  Linguistics, {ACL} 2019, Florence, Italy, July 28- August 2, 2019,
                  Volume 1: Long Papers},
  pages        = {4762--4779},
  publisher    = {Association for Computational Linguistics},
  year         = {2019},
  url          = {https://doi.org/10.18653/v1/p19-1470},
  doi          = {10.18653/V1/P19-1470},
  bibsource    = {dblp computer science bibliography, https://dblp.org}
}

@inproceedings{DBLP:conf/aaai/HwangBBDSBC21,
  author       = {Jena D. Hwang and
                  Chandra Bhagavatula and
                  Ronan Le Bras and
                  Jeff Da and
                  Keisuke Sakaguchi and
                  Antoine Bosselut and
                  Yejin Choi},
  title        = {(Comet-) Atomic 2020: On Symbolic and Neural Commonsense Knowledge
                  Graphs},
  booktitle    = {Thirty-Fifth {AAAI} Conference on Artificial Intelligence, {AAAI}
                  2021, Thirty-Third Conference on Innovative Applications of Artificial
                  Intelligence, {IAAI} 2021, The Eleventh Symposium on Educational Advances
                  in Artificial Intelligence, {EAAI} 2021, Virtual Event, February 2-9,
                  2021},
  pages        = {6384--6392},
  publisher    = {{AAAI} Press},
  year         = {2021},
  url          = {https://doi.org/10.1609/aaai.v35i7.16792},
  doi          = {10.1609/AAAI.V35I7.16792},
  bibsource    = {dblp computer science bibliography, https://dblp.org}
}

@inproceedings{DBLP:conf/aaai/SpeerCH17,
  author       = {Robyn Speer and
                  Joshua Chin and
                  Catherine Havasi},
  editor       = {Satinder Singh and
                  Shaul Markovitch},
  title        = {ConceptNet 5.5: An Open Multilingual Graph of General Knowledge},
  booktitle    = {Proceedings of the Thirty-First {AAAI} Conference on Artificial Intelligence,
                  February 4-9, 2017, San Francisco, California, {USA}},
  pages        = {4444--4451},
  publisher    = {{AAAI} Press},
  year         = {2017},
  url          = {https://doi.org/10.1609/aaai.v31i1.11164},
  doi          = {10.1609/AAAI.V31I1.11164},
  bibsource    = {dblp computer science bibliography, https://dblp.org}
}

@inproceedings{DBLP:conf/acl/Mohammad18,
  author       = {Saif M. Mohammad},
  editor       = {Iryna Gurevych and
                  Yusuke Miyao},
  title        = {Obtaining Reliable Human Ratings of Valence, Arousal, and Dominance
                  for 20, 000 English Words},
  booktitle    = {Proceedings of the 56th Annual Meeting of the Association for Computational
                  Linguistics, {ACL} 2018, Melbourne, Australia, July 15-20, 2018, Volume
                  1: Long Papers},
  pages        = {174--184},
  publisher    = {Association for Computational Linguistics},
  year         = {2018},
  url          = {https://aclanthology.org/P18-1017/},
  doi          = {10.18653/V1/P18-1017},
  bibsource    = {dblp computer science bibliography, https://dblp.org}
}

@inproceedings{DBLP:conf/nips/VaswaniSPUJGKP17,
  author       = {Ashish Vaswani and
                  Noam Shazeer and
                  Niki Parmar and
                  Jakob Uszkoreit and
                  Llion Jones and
                  Aidan N. Gomez and
                  Lukasz Kaiser and
                  Illia Polosukhin},
  editor       = {Isabelle Guyon and
                  Ulrike von Luxburg and
                  Samy Bengio and
                  Hanna M. Wallach and
                  Rob Fergus and
                  S. V. N. Vishwanathan and
                  Roman Garnett},
  title        = {Attention is All you Need},
  booktitle    = {Advances in Neural Information Processing Systems 30: Annual Conference
                  on Neural Information Processing Systems 2017, December 4-9, 2017,
                  Long Beach, CA, {USA}},
  pages        = {5998--6008},
  year         = {2017},
  url          = {https://proceedings.neurips.cc/paper/2017/hash/3f5ee243547dee91fbd053c1c4a845aa-Abstract.html},
  bibsource    = {dblp computer science bibliography, https://dblp.org}
}

@article{DBLP:journals/corr/abs-2603-15031,
  author       = {Kimi Team},
  title        = {Attention Residuals},
  journal      = {CoRR},
  volume       = {abs/2603.15031},
  year         = {2026},
  url          = {https://doi.org/10.48550/arXiv.2603.15031},
  doi          = {10.48550/ARXIV.2603.15031},
  eprinttype   = {arXiv},
  eprint       = {2603.15031},
  bibsource    = {dblp computer science bibliography, https://dblp.org}
}

@article{DBLP:journals/corr/abs-2512-02556,
  author       = {DeepSeek{-}AI},
  title        = {DeepSeek-V3.2: Pushing the Frontier of Open Large Language Models},
  journal      = {CoRR},
  volume       = {abs/2512.02556},
  year         = {2025},
  url          = {https://doi.org/10.48550/arXiv.2512.02556},
  doi          = {10.48550/ARXIV.2512.02556},
  eprinttype   = {arXiv},
  eprint       = {2512.02556},
  bibsource    = {dblp computer science bibliography, https://dblp.org}
}

@inproceedings{DBLP:conf/naacl/LiGBGD16,
  author       = {Jiwei Li and
                  Michel Galley and
                  Chris Brockett and
                  Jianfeng Gao and
                  Bill Dolan},
  editor       = {Kevin Knight and
                  Ani Nenkova and
                  Owen Rambow},
  title        = {A Diversity-Promoting Objective Function for Neural Conversation Models},
  booktitle    = {{NAACL} {HLT} 2016, The 2016 Conference of the North American Chapter
                  of the Association for Computational Linguistics: Human Language Technologies,
                  San Diego California, USA, June 12-17, 2016},
  pages        = {110--119},
  publisher    = {The Association for Computational Linguistics},
  year         = {2016},
  url          = {https://doi.org/10.18653/v1/n16-1014},
  doi          = {10.18653/V1/N16-1014},
  bibsource    = {dblp computer science bibliography, https://dblp.org}
}

@inproceedings{DBLP:conf/www/JiangRMR19,
  author       = {Shaojie Jiang and
                  Pengjie Ren and
                  Christof Monz and
                  Maarten de Rijke},
  editor       = {Ling Liu and
                  Ryen W. White and
                  Amin Mantrach and
                  Fabrizio Silvestri and
                  Julian J. McAuley and
                  Ricardo Baeza{-}Yates and
                  Leila Zia},
  title        = {Improving Neural Response Diversity with Frequency-Aware Cross-Entropy
                  Loss},
  booktitle    = {The World Wide Web Conference, {WWW} 2019, San Francisco, CA, USA,
                  May 13-17, 2019},
  pages        = {2879--2885},
  publisher    = {{ACM}},
  year         = {2019},
  url          = {https://doi.org/10.1145/3308558.3313415},
  doi          = {10.1145/3308558.3313415},
  bibsource    = {dblp computer science bibliography, https://dblp.org}
}

@inproceedings{DBLP:journals/corr/KingmaB14,
  author       = {Diederik P. Kingma and
                  Jimmy Ba},
  editor       = {Yoshua Bengio and
                  Yann LeCun},
  title        = {Adam: {A} Method for Stochastic Optimization},
  booktitle    = {3rd International Conference on Learning Representations, {ICLR} 2015,
                  San Diego, CA, USA, May 7-9, 2015, Conference Track Proceedings},
  year         = {2015},
  url          = {http://arxiv.org/abs/1412.6980},
  bibsource    = {dblp computer science bibliography, https://dblp.org}
}

@inproceedings{DBLP:conf/emnlp/PenningtonSM14,
  author       = {Jeffrey Pennington and
                  Richard Socher and
                  Christopher D. Manning},
  editor       = {Alessandro Moschitti and
                  Bo Pang and
                  Walter Daelemans},
  title        = {Glove: Global Vectors for Word Representation},
  booktitle    = {Proceedings of the 2014 Conference on Empirical Methods in Natural
                  Language Processing, {EMNLP} 2014, October 25-29, 2014, Doha, Qatar,
                  {A} meeting of SIGDAT, a Special Interest Group of the {ACL}},
  pages        = {1532--1543},
  publisher    = {{ACL}},
  year         = {2014},
  url          = {https://doi.org/10.3115/v1/d14-1162},
  doi          = {10.3115/V1/D14-1162},
  bibsource    = {dblp computer science bibliography, https://dblp.org}
}

@inproceedings{DBLP:conf/emnlp/MillerFDKBW16,
  author       = {Alexander H. Miller and
                  Adam Fisch and
                  Jesse Dodge and
                  Amir{-}Hossein Karimi and
                  Antoine Bordes and
                  Jason Weston},
  editor       = {Jian Su and
                  Xavier Carreras and
                  Kevin Duh},
  title        = {Key-Value Memory Networks for Directly Reading Documents},
  booktitle    = {Proceedings of the 2016 Conference on Empirical Methods in Natural
                  Language Processing, {EMNLP} 2016, Austin, Texas, USA, November 1-4,
                  2016},
  pages        = {1400--1409},
  publisher    = {The Association for Computational Linguistics},
  year         = {2016},
  url          = {https://doi.org/10.18653/v1/d16-1147},
  doi          = {10.18653/V1/D16-1147},
  bibsource    = {dblp computer science bibliography, https://dblp.org}
}

@article{fleiss1973equivalence,
  title={The equivalence of weighted kappa and the intraclass correlation coefficient as measures of reliability},
  author={Fleiss, Joseph L and Cohen, Jacob},
  journal={Educational and psychological measurement},
  volume={33},
  number={3},
  pages={613--619},
  year={1973},
  publisher={Sage Publications Sage CA: Thousand Oaks, CA}
}

\appendix

\section{Commonsense Knowledge Generation with COMET}
\label{sec:appendix_comet}

\begin{figure}[t]
    \centering
    \includegraphics[width=\columnwidth]{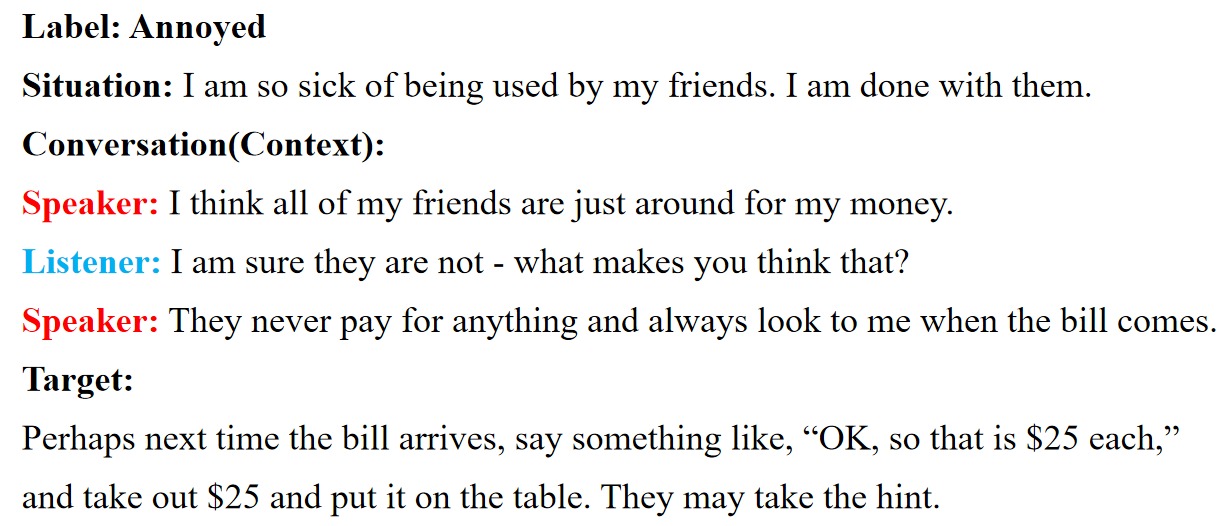}
    \caption{A dialogue example from the EmpatheticDialogues dataset.}
    \label{fig:comet_example_dialogue}
\end{figure}

Figure~\ref{fig:comet_example_dialogue} shows an example from the EmpatheticDialogues dataset. Each instance includes an emotion label, a situation description, the dialogue context, and the target response.

For commonsense generation, we use the BART-based COMET model\footnote{\url{https://github.com/allenai/comet-atomic-2020}} trained on ATOMIC-2020~\cite{DBLP:conf/aaai/HwangBBDSBC21}, applying it to both the dialogue context and the situation description.

\begin{figure}[t]
    \centering
    \includegraphics[width=\columnwidth]{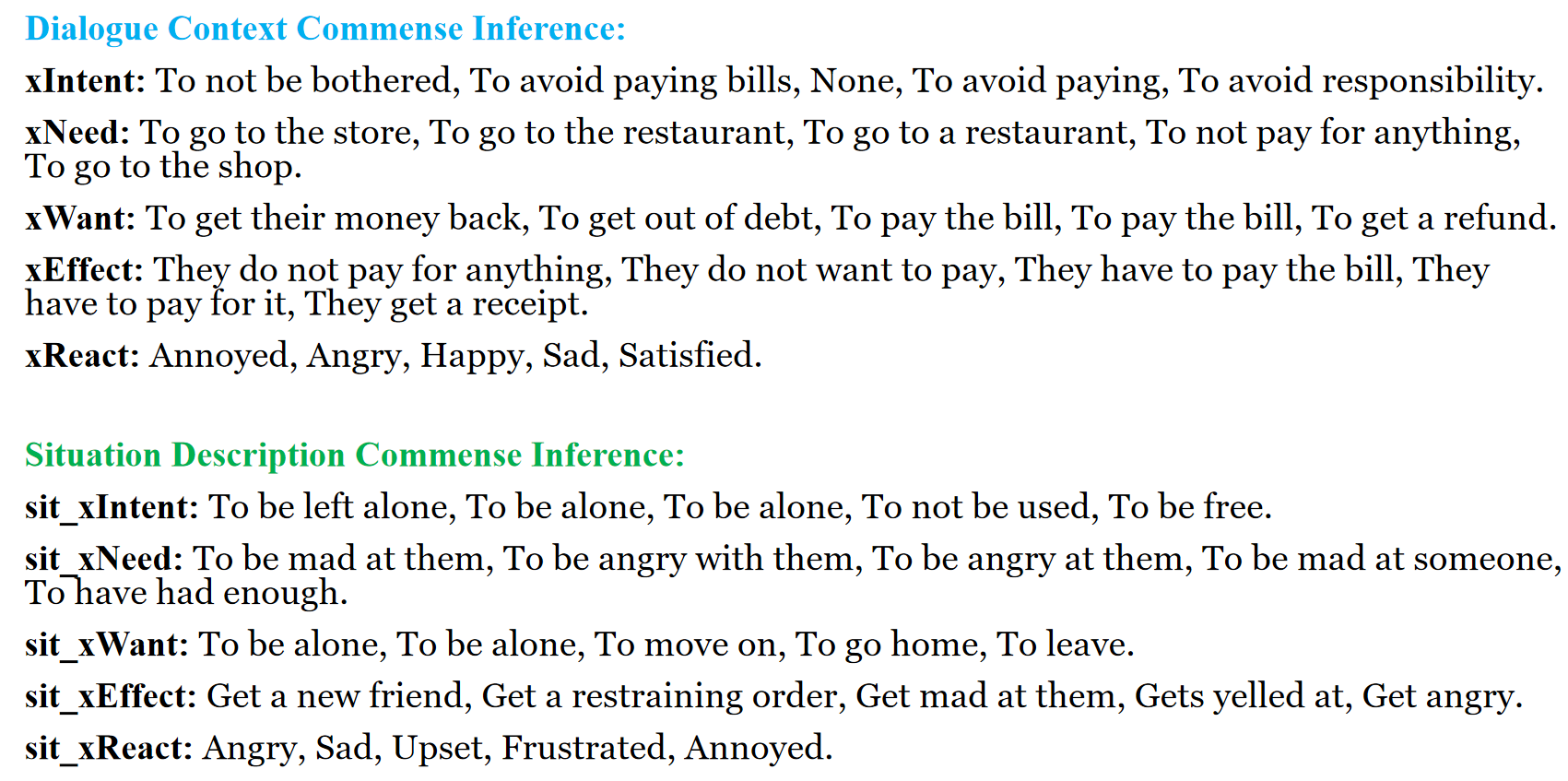}
    \caption{COMET-generated commonsense inferences for both the dialogue context and the corresponding situation description.}
    \label{fig:comet_example_knowledge}
\end{figure}

Figure~\ref{fig:comet_example_knowledge} shows the resulting commonsense inferences. 
Commonsense generated from the dialogue context is denoted using the original COMET relation names (e.g., \textit{xIntent}), whereas commonsense generated from the situation description is prefixed with \textit{sit\_} (e.g., \textit{sit\_xIntent}) for clarity. 
The two commonsense sources provide complementary information: dialogue-context commonsense mainly reflects the ongoing conversational interaction, whereas situation commonsense captures the broader situational background associated with the speaker's emotional state. 
These complementary commonsense representations are subsequently processed by our Dynamic Commonsense Coordination Framework during dialogue understanding and response generation.

We adopt five commonsense relation types, which correspond to both cognitive and affective dimensions of empathy. Specifically, the first four relations capture the user's cognitive state, while \textit{xReact} captures the affective response.

\begin{itemize}
    \item \textbf{xIntent}: PersonX's likely intention before the event.
    \item \textbf{xNeed}: PersonX's likely prerequisite or need before the event.
    \item \textbf{xWant}: PersonX's likely desire or goal after the event.
    \item \textbf{xEffect}: The likely effect of the event on PersonX.
    \item \textbf{xReact}: PersonX's likely emotional reaction after the event.
\end{itemize}

\section{Methodology Details}
\label{sec:appendix_method}

This section expands on several design choices that are summarized more briefly in the main paper.

\subsection{Encoding Design}
\label{sec:encoding_details}

The main paper presents the encoding procedure. Here we briefly explain the motivation behind the two encoding strategies.

The four cognitive relations (\textit{xIntent}, \textit{xNeed}, \textit{xWant}, and \textit{xEffect}) describe high-level event semantics. Therefore, we use the \texttt{[CLS]} representation as a sequence-level summary after Transformer encoding.

In contrast, \textit{xReact} usually consists of short affective expressions whose emotional information may be distributed across multiple tokens. We therefore apply mean pooling to aggregate token-level representations into a more stable affective representation. 
This follows the affective commonsense encoding used in CEM~\cite{DBLP:conf/aaai/SabourZH22}.

\subsection{Attention-Residual Fusion Details}
\label{sec:attention_residual}

Unlike conventional residual connections that directly add the cross-attention output to the original representation, our attention-residual mechanism treats the dialogue representation, the situation representation, and the cross-attention output as three candidate sources.

The scores are content-dependent and computed from each source vector after LayerNorm via a learnable projection. The normalization operation serves a similar purpose to RMSNorm in the original Attention Residuals formulation, preventing sources with larger representation magnitudes from dominating the attention weights.

The softmax normalization introduces competition among the three sources, encouraging the model to emphasize the most relevant source(s) for the current context while retaining the original representations as direct fusion candidates.

This design is inspired by recent work on attention residuals~\cite{DBLP:journals/corr/abs-2603-15031}, which replaces fixed residual accumulation with learnable attention over depth. The key insight is that selective, content-dependent aggregation can also be applied to choosing among different information sources. We adapt this idea from the depth dimension to the source dimension: instead of simply adding the cross-attention output to the original representation, we allow the model to learn how to balance multiple sources.

The mathematical formulation is provided in Section~\ref{sec:sce_attnres}.

\subsection{AGCF Implementation Details}
\label{sec:appendix_agcf}

As described in Section~\ref{sec:agcf}, AGCF processes the dialogue memory $\mathbf{M}^{ctx}$ and situation memory $\mathbf{M}^{sit}$ independently. In the implementation, separate linear scoring heads are used for the two memories: $\mathbf{W}_s^{ctx}$ for the dialogue memory and $\mathbf{W}_s^{sit}$ for the situation memory. This design allows the model to learn different relevance criteria for each commonsense source.

Because the softmax weights sum to one, directly applying them would reduce the average relation weight to $1/|R|$. We therefore multiply the normalized weights by $|R|=5$, making their average equal to one and helping preserve the overall scale of the relation memories.

From a broader perspective, this filtering mechanism is related to the memory addressing and reading process in Key-Value Memory Networks~\cite{DBLP:conf/emnlp/MillerFDKBW16}, where a query addresses a set of memory slots through relevance scores and retrieves the corresponding memory contents. 
In AGCF, the global dialogue representation $\mathbf{q}$ acts as the query, while each relation vector $\bar{\mathbf{m}}_r$ is used as both the key and value. 
The resulting attention weights $\alpha_r$ control each relation's contribution to the filtered memory.
Unlike Key-Value Memory Networks, which operate on external structured knowledge, AGCF filters input-dependent commonsense representations and produces a reweighted version of the same memory rather than returning retrieved facts.

\subsection{ICAD Implementation Details}
\label{sec:appendix_icad}

As described in Section~\ref{sec:icad}, ICAD performs commonsense retrieval at every decoding step. 
For a given input, the filtered relation memories $\tilde{\mathbf{M}}^{ctx}$ and $\tilde{\mathbf{M}}^{sit}$ are computed once and remain fixed throughout autoregressive decoding.
 
What changes across decoding steps is the decoder hidden state $\mathbf{h}_t^{dec}$, which serves as the cross-attention query, and consequently the attention distribution over the relation memories.
This allows the decoder to retrieve different commonsense relations as the semantic focus of the generated response changes.

\subsection{Frequency-Aware Weight Computation}
\label{sec:appendix_face}

Following CEM~\cite{DBLP:conf/aaai/SabourZH22}, we dynamically accumulate token frequencies from the model predictions produced during training. Let $f_i$ denote the accumulated predicted frequency of token $c_i$. Its relative frequency is

\begin{equation}
RF_i
=
\frac{f_i}
{\sum_{j=1}^{|V|} f_j}.
\end{equation}

The unnormalized frequency-aware weight is computed as

\begin{equation}
\hat{w}_i
=
a \cdot RF_i + 1,
\end{equation}

where

\begin{equation}
a
=
-\frac{1}
{\max_j RF_j}.
\end{equation}

The weights are subsequently normalized so that their sum equals the vocabulary size:

\begin{equation}
w_i
=
\frac{|V| \cdot \hat{w}_i}
{\sum_{j=1}^{|V|}\hat{w}_j}.
\end{equation}

Consequently, frequently predicted tokens receive smaller weights, whereas less frequent tokens receive larger weights. The resulting weights are used in the FACE loss~\cite{DBLP:conf/www/JiangRMR19} defined in Section~\ref{sec:training}.

\section{Experimental Settings}
\label{sec:appendix_experimental_settings}

We describe the full experimental configuration here. All experiments are implemented using PyTorch\footnote{\url{https://pytorch.org/}} and run on NVIDIA RTX 4090 GPUs. For a fair comparison, we reproduce the CEM baseline~\cite{DBLP:conf/aaai/SabourZH22} from its official implementation\footnote{\url{https://github.com/Sahandfer/CEM}} and keep its hyperparameters unchanged.

\subsection{Dataset}

We evaluate our framework on the Empathetic-Dialogues dataset~\cite{DBLP:conf/acl/RashkinSLB19}, a standard benchmark for empathetic response generation.
The dataset contains 24,850 multi-turn dialogues grounded in 32 emotion categories. 
Following the standard split, we use 19,533 dialogues for training, 2,770 for validation, and 2,547 for testing.

\subsection{Hyperparameters}

We adopt the same hyperparameters as the original CEM implementation~\cite{DBLP:conf/aaai/SabourZH22}. The key settings are as follows.

\begin{itemize}
    \item \textbf{Optimizer:} Adam~\cite{DBLP:journals/corr/KingmaB14}.
    \item \textbf{Learning rate:} The initial learning rate is set to $1 \times 10^{-4}$.
    \item \textbf{Batch size:} The batch size during training is $16$.
    \item \textbf{Hidden dimension:} The hidden dimension for all components is set to $300$.
    \item \textbf{Embedding dimension:} The word embedding dimension is $300$, initialized with pre-trained GloVe vectors~\cite{DBLP:conf/emnlp/PenningtonSM14}.
    \item \textbf{Gradient clipping:} The maximum gradient norm is set to $2.0$.
    \item \textbf{Beam size:} Beam search with a beam size of $5$ is used during decoding.
    \item \textbf{Random seed:} The random seed is set to $42$ for reproducibility.
    \item \textbf{Transformer:} Both the encoder and decoder contain $1$ layer with $2$ attention heads. The key depth is $40$, and the intermediate feed-forward dimension is $50$.
\end{itemize}

\subsection{Training Details}

We use the Adam~\cite{DBLP:journals/corr/KingmaB14} and apply early stopping with a patience of 3 evaluations. 
The validation set is evaluated every 2,000 iterations, and training stops if the validation perplexity fails to improve for three consecutive checks. 
For each model, we report results from the best-performing validation checkpoint.

\section{Comparison with Recent Association-based Models}
\label{sec:comparison_recent_models}

DCC, IAMM~\cite{DBLP:conf/acl/YangRYSCZL24}, and SDAM~\cite{DBLP:journals/ipm/YangRWSZL24} all exploit information beyond the immediate
dialogue representation, but differ in their modeling objectives,
granularities, and use of situation and commonsense information.
Table~\ref{tab:mechanism_comparison} summarizes their main differences.

\begin{table*}[t]
\centering
\caption{Comparison of modeling mechanisms across DCC, IAMM, and SDAM.}
\label{tab:mechanism_comparison}
\setlength{\tabcolsep}{4.5pt}
\renewcommand{\arraystretch}{1.15}
\small
\begin{tabular}{p{1.4cm}p{3.1cm}p{3.0cm}p{3.2cm}p{4.0cm}}
\toprule
\textbf{Model}
& \textbf{Primary objective}
& \textbf{Modeling granularity}
& \textbf{Use of situation}
& \textbf{Dynamic mechanism} \\
\midrule

SDAM
& Associate situations with dialogues
& Explicit lexical and implicit knowledge-based associations
& Treats the situation as broader context and models its associations
with the dialogue
& Selects mutually relevant keywords through bidirectional filtering
and captures implicit associations with a reasoning knowledge-based
hypergraph network \\
\midrule

IAMM
& Capture associated information across dialogue utterances
& Word- and utterance-level associations
& Iteratively associates each utterance with the situation, dialogue
history, and associative memory
& Uses second-order interaction attention to identify associated words
and progressively update the memory \\
\midrule

DCC
& Coordinate commonsense across understanding and generation
& Source-, relation-, and token-level commonsense coordination
& Constructs separate dialogue- and situation-derived commonsense
representations
& Performs cross-source coordination with SCE-AttnRes, relation-level
filtering with AGCF, and token-dependent retrieval with ICAD \\

\bottomrule
\end{tabular}
\end{table*}

\paragraph{Situation modeling.}
Situation modeling differs across the three approaches.
SDAM directly targets associations between the situation and dialogue.
It models explicit associations by selecting mutually relevant
keywords and captures more complex implicit associations through a
reasoning knowledge-based hypergraph. 
IAMM instead uses the situation
as one source of information in an iterative comprehension process.
For each dialogue utterance, it identifies associated information from
the situation, dialogue history, and previously constructed memory.
DCC maintains separate commonsense representations derived from the
dialogue and situation. SCE-AttnRes then coordinates the two sources
while retaining their source-specific information.

\paragraph{Commonsense modeling.}
The three approaches also differ in how they use reasoning knowledge. 
SDAM employs reasoning knowledge to construct implicit situation--dialogue
associations, while IAMM treats commonsense inferences as implicit
information from which associated words are iteratively identified.
DCC organizes COMET inferences into five predefined relations:
\textit{xIntent}, \textit{xNeed}, \textit{xWant}, \textit{xEffect},
and \textit{xReact}. AGCF independently estimates the relevance of
these relations for dialogue- and situation-derived memories, thereby
filtering commonsense before response generation.

\paragraph{Dynamic processing.}
The most notable difference is where dynamic processing occurs.
IAMM dynamically updates an associative memory while processing
successive utterances, whereas SDAM mainly establishes explicit and
implicit associations during dialogue understanding. 
DCC introduces dynamic processing at multiple stages. SCE-AttnRes coordinates the two
commonsense sources, AGCF performs input-dependent relation weighting,
and ICAD retrieves relation-level commonsense according to the evolving
decoder state. Thus, ICAD can change its relation preference as the
response unfolds.

\paragraph{Performance and scope.}
IAMM and SDAM achieve stronger overall automatic results than DCC, particularly in emotion accuracy and response diversity, reflecting the strength of their fine-grained associative modeling. 
DCC instead offers a complementary investigation of how structured commonsense can be coordinated across sources, relations, and decoding steps. 
Because DCC is built on the official CEM codebase, its comparison with CEM more directly isolates the contribution of the proposed coordination mechanisms. The ablation results further examine the individual effects of SCE-AttnRes, AGCF, and ICAD.

\paragraph{Limitations.}
A limitation of DCC is that it does not discover arbitrary word-level associations. Its commonsense coordination is restricted to five predefined COMET~\cite{DBLP:conf/acl/BosselutRSMCC19,DBLP:conf/aaai/HwangBBDSBC21} relations, which may omit relevant associations outside this set. Combining relation-level coordination with more flexible word-level association or memory updating remains a promising direction.

\section{Additional Ablation Details}
\label{sec:ablation_details}

We evaluate three ablation variants. Removing SCE-AttnRes disables cross-source commonsense interaction during dialogue understanding, while keeping AGCF and ICAD intact. The variant without AGCF passes unweighted relation memories directly to the decoder. In the ICAD ablation, token-dependent retrieval is replaced with static mean aggregation of the AGCF-filtered memories, with the decoder-side fusion layer unchanged.

Removing SCE-AttnRes decreases emotion accuracy from 46.09\% to
36.59\%, a drop of 9.50 percentage points. It also increases PPL from
36.13 to 36.89 and reduces Dist-1 and Dist-2 by 0.26 and 1.05,
respectively. The consistent degradations across all metrics confirm that cross-source coordination benefits both emotion discrimination and generation diversity.

Removing AGCF results in the highest PPL (37.72) and the lowest
Dist-1/2 scores (0.73/3.66), indicating that AGCF primarily serves to filter out less useful commonsense before decoding. 
Acc increases slightly from 46.09\% to 46.89\%.  
This modest Acc increase may be due to the additional xReact signals retained without filtering, which
provide supplementary emotional cues for classification-though at the
cost of more noisy information reaching the decoder.

Replacing ICAD with static aggregation increases PPL from 36.13 to
37.48 and decreases Acc from 46.09\% to 44.99\%, while increasing
Dist-1/2 from 1.03/4.93 to 1.05/5.59. Thus, ICAD does not uniformly
increase surface diversity; instead, it conditions commonsense use on
the evolving decoder state, improving likelihood and emotion
discrimination while constraining lexical variation.

\section{AGCF Relation-Weight Visualization}
\label{sec:agcf_visualization}

We analyze AGCF's relation weights on all 5,255 test instances (derived from the 2,547 test dialogues) to see how it differentiates among commonsense relations. 
AGCF uses separate scoring heads for dialogue- and situation-derived memories, so we analyze the two weight distributions independently. 
These weights are computed from the dialogue, situation, and their commonsense memories only; the target response does not influence AGCF scoring.

The softmax weights are rescaled by the number of relations,
$|\mathcal{R}|=5$. Consequently, the five weights sum to 5 and have a
mean of 1. The dashed line in each figure denotes this uniform-weighting
reference: values above 1 indicate emphasized relations, whereas values
below 1 indicate suppressed relations.

\subsection{Case 3520: Different Preferences across Sources}

Case 3520 concerns a parent's anxiety about a son starting college. Its
situation is ``anxious awaiting son to start college.'' The dialogue
provides a more detailed progression from anxiety to acceptance,
including statements such as ``I just want the best for him,''
``let him grow up,'' and ``lose the control freak in me.''

\begin{figure*}[t]
\centering
\includegraphics[width=\textwidth]
{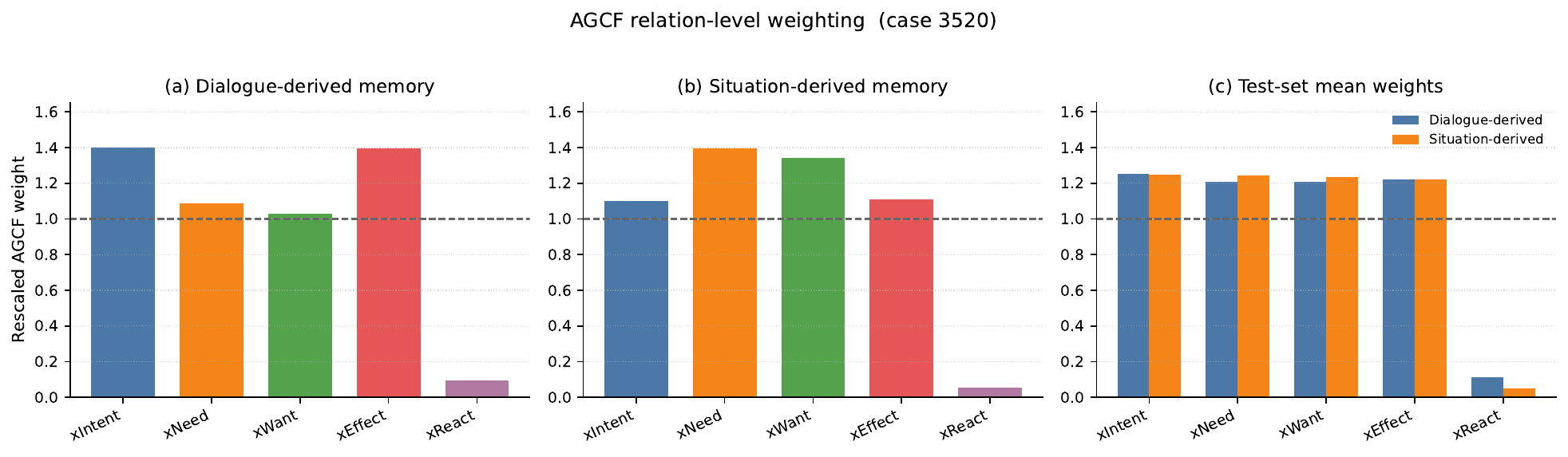}
\caption{AGCF relation weights for Case 3520 and the corresponding
test-set means. Dialogue- and situation-derived memories are scored by
separate AGCF heads. The dashed line denotes the uniform-weighting
reference value of 1 after rescaling.}
\label{fig:agcf_case_3520}
\end{figure*}

As shown in Figure~\ref{fig:agcf_case_3520}, the dialogue-derived head
assigns its largest weights to \textit{xIntent} (1.400) and
\textit{xEffect} (1.393). This preference is consistent with the
intention- and outcome-related content in the extended dialogue,
including relinquishing control, allowing the son to grow, and
expecting that he will be fine. In contrast, the situation-derived head
emphasizes \textit{xNeed} (1.395) and \textit{xWant} (1.340), whose
COMET inferences concern preparation, support, and desired actions
associated with starting college. Thus, the two scoring heads produce
different relation distributions for the same underlying event.

\subsection{Case 1724: Outcome-Oriented Situation Weighting}

Case 1724 describes a speaker whose daughter unexpectedly went into
labor in another state while the speaker could not be present. The
dialogue additionally provides the eventual outcome: everyone is safe,
and the initially distressing event becomes joyous.

\begin{figure*}[t]
\centering
\includegraphics[width=\textwidth]
{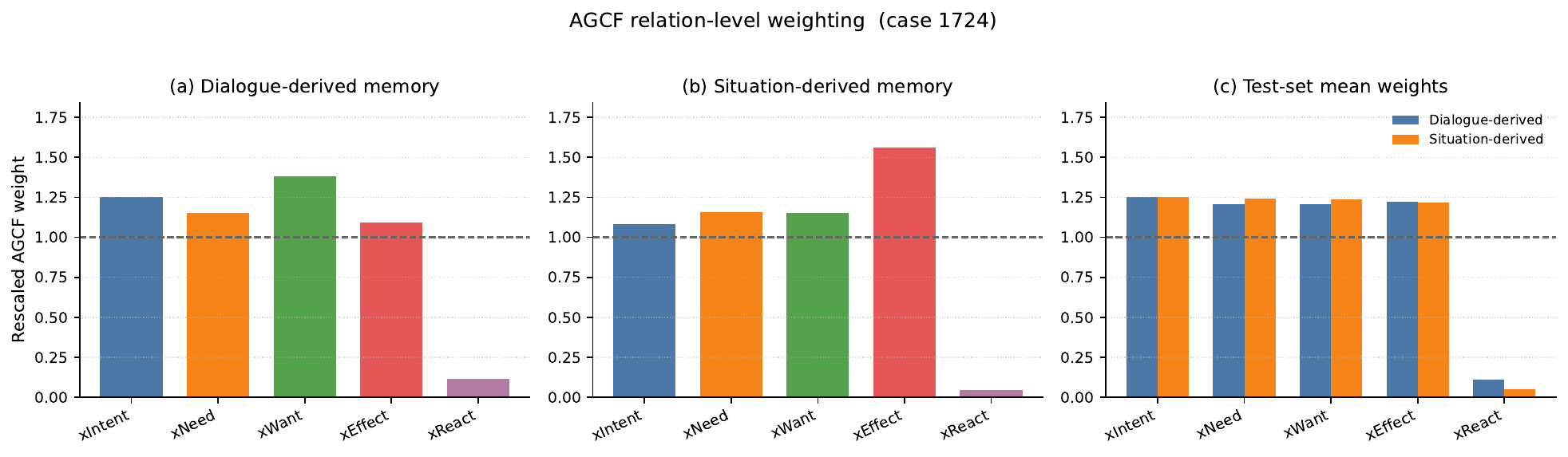}
\caption{AGCF relation weights for Case 1724 and the corresponding
test-set means. The dialogue contains both the initial concern and the
positive outcome, whereas the situation concentrates on the premature
labor event. The dashed line denotes the uniform-weighting reference
value of 1.}
\label{fig:agcf_case_1724}
\end{figure*}

As shown in Figure~\ref{fig:agcf_case_1724}, the dialogue-derived head
assigns its largest weight to \textit{xWant} (1.383), followed by
\textit{xIntent} (1.254). The corresponding COMET inferences include
desired states such as having the baby, returning home, and being
happy. By contrast, the situation-derived head strongly emphasizes
\textit{xEffect} (1.561), reflecting the immediate consequences of the
labor event and the speaker's inability to be present. 
This case
therefore exhibits a different source-specific pattern from Case 3520:
the dominant relations change from
\textit{xIntent}/\textit{xNeed} to
\textit{xWant}/\textit{xEffect}.

\subsection{Test-Set Distribution}

Across the test set, the mean dialogue-derived weights for
\textit{xIntent}, \textit{xNeed}, \textit{xWant}, \textit{xEffect},
and \textit{xReact} are 1.252, 1.207, 1.208, 1.222, and 0.112,
respectively. The corresponding situation-derived means are 1.250,
1.243, 1.236, 1.219, and 0.053. Thus, in the decoder-side relation
memories, AGCF generally emphasizes cognitive commonsense and assigns
substantially lower weights to \textit{xReact}. This observation should
not be interpreted as showing that affective information is globally
unimportant, because emotion-related signals can still enter DCC
through the dialogue encoder, SCE-AttnRes, and emotion-classification
pathway.

The case-level results further show that AGCF does not apply an
identical relation distribution to every input. 
It emphasizes
\textit{xIntent}/\textit{xEffect} versus
\textit{xNeed}/\textit{xWant} in Case 3520, and
\textit{xWant} versus \textit{xEffect} in Case 1724. 
These differences suggest that relation-level filtering varies with both input content and commonsense source. 
Together with the ablation results, this analysis provides complementary evidence that AGCF changes the relation-level information supplied to the decoder. 
The case identifiers correspond to the original test-set ordering, so the full inputs and target responses can be retrieved from the dataset.

\section{ICAD Token-Level Attention Visualization}
\label{sec:icad_visualization}

We record ICAD's token-level attention distributions on representative test instances to see whether it dynamically adjusts commonsense retrieval during generation. 
The two memory sources use separate attention heads; probabilities are averaged across heads and visualized independently for each source.

Different from the rescaled AGCF weights, ICAD uses standard softmax attention.
For each generated token, the weights over the five COMET relations sum
to 1. 
The horizontal axis of each heatmap represents greedily generated
tokens, while the vertical axis represents the five relations. 
The attention associated with a token is computed from the preceding
decoder state; neither the target response nor future generated tokens
are used. The responses shown in the case studies are generated by
greedy decoding and do not necessarily match the ground-truth target
responses.

\subsection{Case 3550: From Affective Reaction to Event Information}

Case 3550 concerns a dangerous encounter with a police vehicle. The
dialogue states that police officers should drive safely even when
responding to a call and describes the speaker reversing to avoid the
vehicle. The situation provides further details: a squad car entered
the speaker's lane head-on while its driver was looking away, leaving
the speaker with nowhere else to go. Given this input, DCC generates:

\begin{quote}
\small
\textbf{Generated response:} ``That is terrible! Did you call the
cops?''
\end{quote}

\begin{figure*}[t]
\centering
\includegraphics[width=0.85\textwidth]
{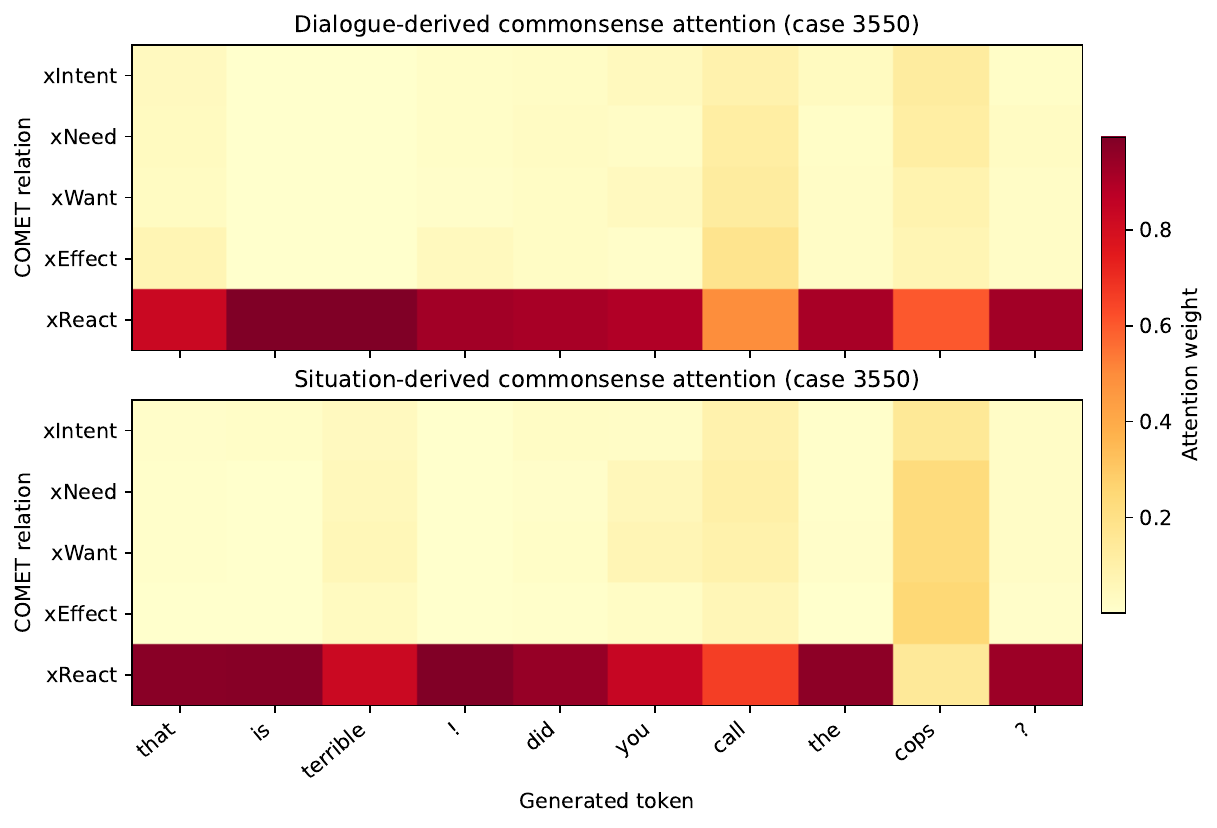}
\caption{Token-level ICAD attention for Case 3550. The upper and lower
heatmaps show attention over dialogue- and situation-derived
commonsense memories, respectively. Each column is normalized over the
five COMET relations.}
\label{fig:icad_case_3550}
\end{figure*}

As shown in Figure~\ref{fig:icad_case_3550}, both memory branches
strongly emphasize \textit{xReact} while generating the affective
phrase ``that is terrible.'' In particular, the dialogue-derived
attention on \textit{xReact} reaches approximately 0.99 for
``terrible,'' while the situation-derived branch assigns it 0.82. This
preference is semantically consistent with the reaction inferences,
which include ``scared,'' ``angry,'' ``annoyed,'' and ``nervous.''

The attention pattern changes when the response proceeds to more
event-specific content. When generating ``cops,'' the
situation-derived attention on \textit{xReact} decreases to 0.15,
whereas \textit{xEffect}, \textit{xNeed}, and \textit{xWant} receive
0.25, 0.23, and 0.23, respectively. These relations contain
police- and event-related inferences such as ``a police officer,''
``a cop,'' and ``get in trouble.'' By contrast, the dialogue-derived
branch remains primarily focused on \textit{xReact} (0.60). Thus, the
two branches provide complementary retrieval patterns: the dialogue
memory retains the affective interpretation, while the situation memory
shifts toward information associated with the event and its possible
consequences.

\subsection{Case 250: From Fear to a Concrete Experience}

In Case 250, the dialogue and situation contain the same statement:

\begin{quote}
\small
``I watch so many scary movies that I have not been able to sleep alone
in years due to fear.''
\end{quote}

Given this input, DCC generates:

\begin{quote}
\small
\textbf{Generated response:} ``What movie did you see?''
\end{quote}

\begin{figure*}[t]
\centering
\includegraphics[width=0.85\textwidth]
{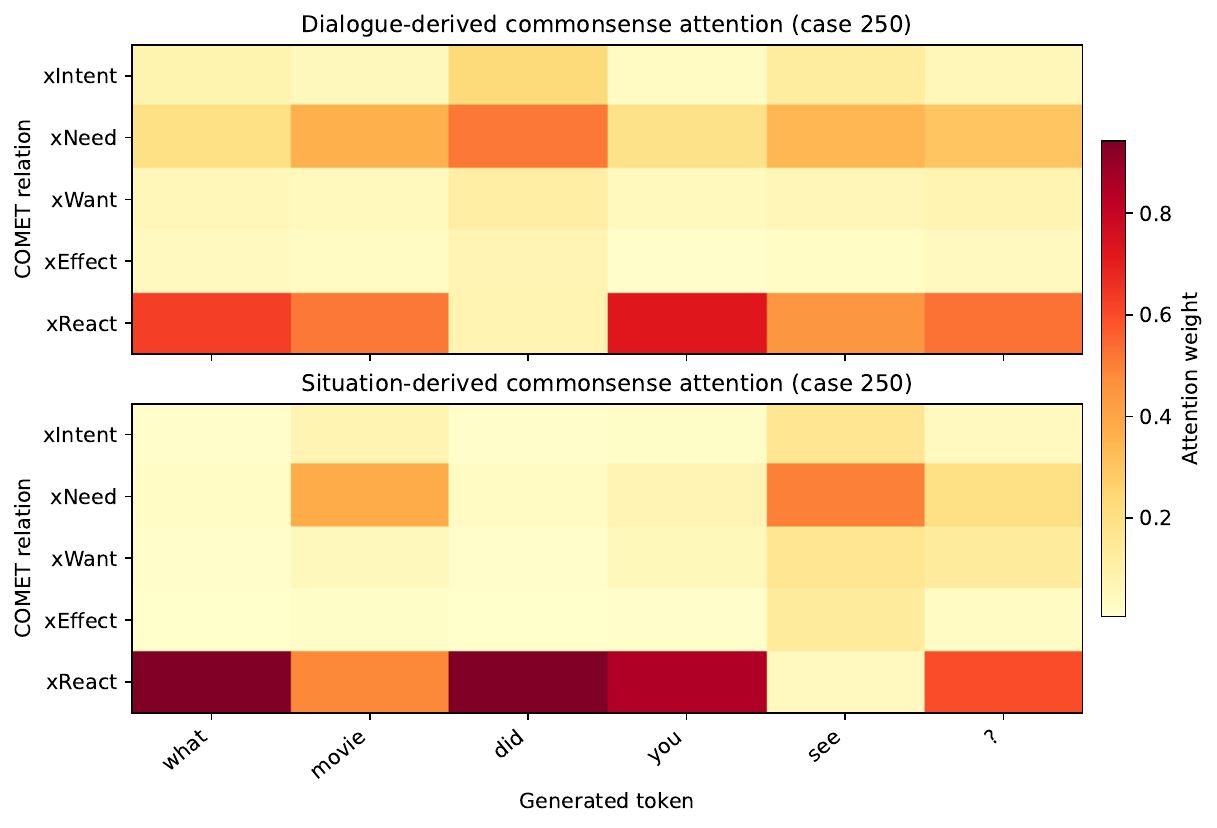}
\caption{Token-level ICAD attention for Case 250. Although the dialogue
and situation contain the same text, their independently processed
commonsense memories exhibit different retrieval trajectories during
generation. Each column sums to 1.}
\label{fig:icad_case_250}
\end{figure*}

As shown in Figure~\ref{fig:icad_case_250}, the response initially
relies heavily on affective commonsense. For ``what,'' the
dialogue- and situation-derived branches assign 0.62 and 0.94 to
\textit{xReact}, respectively. The corresponding inferences include
``scared,'' ``terrified,'' ``frightened,'' and ``nervous,'' which
match the speaker's stated fear.

The two branches subsequently follow different trajectories. When
generating ``did,'' the dialogue-derived branch shifts to
\textit{xNeed} (0.51), whereas the situation-derived branch remains
focused on \textit{xReact} (0.94). For ``you,'' the dialogue-derived
branch returns to \textit{xReact} (0.72). When generating the
content word ``see,'' however, the situation-derived branch shifts
sharply from \textit{xReact} to \textit{xNeed}: its
\textit{xReact} weight decreases to 0.04, while
\textit{xNeed} increases to 0.50. The corresponding
\textit{xNeed} inferences include ``to go to the movies,'' ``to watch
scary movies,'' and ``to go to the theater,'' providing a semantic
connection to the generated question.

Although the dialogue and situation texts are identical in this case,
their attention distributions are not. This difference arises because
the two commonsense memories are processed by independent AGCF scoring
heads and retrieved through separate ICAD attention heads. The case
therefore illustrates that source-specific processing can produce
different decoding behavior even when the two textual inputs overlap.

\subsection{Discussion}

The two cases exhibit complementary forms of dynamic retrieval. In
Case 3550, ICAD moves from an affective reaction toward event-related
information as the response unfolds, while maintaining different
preferences across the two memory sources. In Case 250, the
situation-derived branch shifts from \textit{xReact} to
\textit{xNeed} when generating a content word associated with the
speaker's experience.

These visualizations show that ICAD attention is neither constant
across decoding steps nor identical across commonsense sources.
However, attention variation alone does not establish a causal
improvement in response quality. Instead, it provides mechanism-level
evidence that ICAD performs the intended token-dependent retrieval.

The ablation results reveal a trade-off between likelihood-based performance and lexical diversity. Taken together with these visualizations, the evidence suggests that ICAD dynamically constrains the commonsense information used during generation.

\section{LLM-Based Evaluation}
\label{sec:llm_evaluation}

We also conduct a blind pairwise LLM-based comparison between DCC and our reproduced CEM baseline, complementing the automatic evaluation with a more direct assessment of response-level qualities. 
This evaluation examines response-level qualities that are not fully
captured by perplexity, lexical diversity, or emotion classification
accuracy.

\subsection{Evaluation Setup}

We sample 500 instances from the Empathetic-Dialogues test set with a fixed random seed of 2026. 
For each instance, the evaluator is given the situation description, the complete dialogue context, and two candidate responses generated by DCC and CEM. The responses are anonymized as Response A and Response B, and their model identities are not disclosed to the evaluator.

The evaluator is DeepSeek-V3.2~\cite{DBLP:journals/corr/abs-2512-02556}, accessed via the SiliconFlow API\footnote{\url{https://www.siliconflow.com/}} in July 2026. 
The decoding temperature is set to 0, and the maximum output length is 350 tokens. 
The evaluator returns a structured JSON object containing its preference for each criterion and a one-sentence justification. 
Reference responses are not shown to the evaluator because empathetic dialogue permits multiple appropriate responses, and exposing a single reference could introduce reference-expression bias.

The CEM responses used here are generated by our reproduced CEM model
under the same greedy decoding setting as DCC. 
Therefore, this experiment compares DCC with our reproduced CEM generations rather than with outputs reported in the original CEM paper.

\subsection{Evaluation Criteria}

Five criteria are used for comparison:

\begin{itemize}
    \item \textbf{Empathy}: whether the response recognizes, validates,
    or appropriately responds to the speaker's emotional state;

    \item \textbf{Relevance}: whether the response is specific and
    relevant to the given situation and dialogue context;

    \item \textbf{Coherence}: whether the response is fluent, natural,
    and logically consistent as the next dialogue turn;

    \item \textbf{Informativeness}: whether the response contributes
    meaningful content rather than only providing a generic reaction;

    \item \textbf{Overall}: the overall suitability of the response as
    an empathetic continuation of the dialogue.
\end{itemize}

For each criterion, the evaluator selects Response A, Response B, or a
tie. The evaluator is explicitly instructed not to prefer a response
solely because it is longer.

\subsection{Order-Bias Control}

LLM evaluators can be sensitive to presentation order. We therefore evaluate each instance twice, reversing the positions of the two responses in the second round. 
In the first evaluation, DCC is presented as Response A and CEM
as Response B. In the second evaluation, their positions are reversed.

The two judgments are subsequently mapped back to the corresponding
model identities. A model is assigned a win only when both presentation
orders produce the same canonical preference for that model. If the
two judgments disagree, or if only one ordering produces a tie, the
final decision is conservatively recorded as a tie. Formally, for
criterion $c$, the final decision is

\begin{equation}
d_c =
\begin{cases}
\mathrm{DCC}, &
d_c^{(1)} = d_c^{(2)} = \mathrm{DCC}, \\
\mathrm{CEM}, &
d_c^{(1)} = d_c^{(2)} = \mathrm{CEM}, \\
\mathrm{Tie}, & \text{otherwise},
\end{cases}
\end{equation}

where $d_c^{(1)}$ and $d_c^{(2)}$ denote the judgments after mapping the
two response orders back to their model identities. This procedure
results in 1,000 evaluator calls for the 500 sampled instances.

\subsection{Evaluator Prompt}
\label{sec:llm_evaluator_prompt}

The following system prompt is used for all evaluations:

\begin{quote}
\small
\ttfamily
You are an impartial expert evaluator of empathetic dialogue systems.
Given a situation, a dialogue context, and two candidate next
responses, compare the responses independently on the following
criteria:

1. empathy: recognizes, validates, or appropriately responds to the
speaker's emotion;

2. relevance: is specific and relevant to the situation and dialogue
context;

3. coherence: is natural, logically consistent, and fluent as the next
turn;

4. informativeness: contributes meaningful content rather than only a
generic reaction;

5. overall: overall quality as an empathetic next response.

For each criterion, choose exactly ``A'', ``B'', or ``Tie''. Do not
prefer a response merely because it is longer. Do not assume there is
only one correct response. Judge only the displayed text and ignore any
possible identity of the systems. Return JSON only, with exactly these
keys: empathy, relevance, coherence, informativeness, overall,
brief\_reason. The brief\_reason must be one concise sentence and must
not identify systems.
\end{quote}

For each instance, the following user-message template is appended:

\begin{quote}
\small
\ttfamily
Situation:

[\textnormal{SITUATION}]

Dialogue context:

[\textnormal{DIALOGUE CONTEXT}]

Response A:

[\textnormal{RESPONSE A}]

Response B:

[\textnormal{RESPONSE B}]

Return the required JSON object only.
\end{quote}

The required output format is illustrated below:

\begin{verbatim}
{
  "empathy": "A",
  "relevance": "B",
  "coherence": "Tie",
  "informativeness": "A",
  "overall": "A",
  "brief_reason": "A concise justification."
}
\end{verbatim}

\subsection{Results}

Table~\ref{tab:llm_evaluation} shows the final results after double-order consistency checking. All percentages are based on the full set of 500 evaluated instances.

\begin{table}[t]
\centering
\caption{Blind pairwise LLM evaluation of DCC and our reproduced CEM
baseline. Each instance is evaluated under both response orders. A
disagreement between the two orders is conservatively counted as a
tie.}
\label{tab:llm_evaluation}
\setlength{\tabcolsep}{5pt}
\renewcommand{\arraystretch}{1.10}
\small
\begin{tabular}{lccc}
\toprule
\textbf{Criterion}
& \textbf{DCC Win}
& \textbf{Tie}
& \textbf{CEM Win} \\
\midrule
Empathy
& 25.0
& 46.2
& 28.8 \\
Relevance
& 29.8
& 52.0
& 18.2 \\
Coherence
& 28.8
& 50.4
& 20.8 \\
Informativeness
& 20.4
& 66.4
& 13.2 \\
Overall
& 33.0
& 38.0
& 29.0 \\
\bottomrule
\end{tabular}
\end{table}

DCC achieves higher win rates than CEM in relevance, coherence,
informativeness, and overall response quality. 
The largest gap is in relevance: DCC wins 29.8\% of the instances versus CEM's 18.2\%, an 11.6-point advantage. DCC also outperforms CEM by 8.0 points in coherence and 7.2 points in informativeness. These patterns suggest that coordinating dialogue- and situation-derived commonsense improves contextual specificity, logical consistency, and meaningful content.

DCC's overall win rate (33.0\%) is moderately higher than CEM's (29.0\%), though the narrower margin suggests that individual improvements do not always yield a decisive overall preference. Empathy shows a slightly different pattern: CEM edges ahead (28.8\% vs. 25.0\%), but the gap is small and, as reported below, not statistically significant. This does not conflict with DCC's higher emotion classification accuracy, since classification and response-level empathy measure different things—the former is label prediction, the latter is generated expression.

Tie rates are relatively high across all criteria, partly due to the conservative double-order rule and partly because some responses are similarly short. Informativeness has the highest tie rate (66.4\%), indicating that many responses from the two systems convey comparable information. By counting inconsistent order-dependent judgments as ties, we avoid assigning unstable preferences to either system.

\subsection{Statistical Significance}

For each criterion, we conduct a two-sided exact binomial test over
non-tied decisions, with the null hypothesis that DCC and CEM have equal
probability of winning. Because five criteria are tested
simultaneously, we apply the Holm correction to control the family-wise
error rate. Table~\ref{tab:llm_significance} reports the number of
decisive comparisons, DCC's share of these comparisons, the unadjusted
$p$-value, and the Holm-adjusted $p$-value.

\begin{table}[t]
\centering
\caption{Statistical significance of the LLM-based pairwise
evaluation. DCC share is calculated over non-tied decisions.
$^\ast$ denotes significance after Holm correction at $p<0.05$, and
$^{\ast\ast}$ denotes significance at $p<0.01$.}
\label{tab:llm_significance}
\setlength{\tabcolsep}{4pt}
\renewcommand{\arraystretch}{1.10}
\small
\begin{tabular}{lrrrr}
\toprule
\textbf{Criterion}
& \textbf{Decisive}
& \textbf{DCC Share}
& \textbf{$p$}
& \textbf{Holm-$p$} \\
\midrule
Empathy
& 269 & 46.47 & .272 & .545 \\
Relevance
& 240 & 62.08 & $<$.001 & .001$^{\ast\ast}$ \\
Coherence
& 248 & 58.06 & .013 & .039$^\ast$ \\
Informativeness
& 168 & 60.71 & .007 & .027$^\ast$ \\
Overall
& 310 & 53.23 & .281 & .545 \\
\bottomrule
\end{tabular}
\end{table}

DCC's advantages in relevance, coherence, and informativeness remain significant after Holm correction. The empathy disadvantage and the overall win advantage are both non-significant. The results therefore point to targeted improvements in contextual relevance, coherence, and informativeness, not a broad across-the-board gain in empathetic response quality.

\subsection{Limitations of the LLM-Based Evaluation}

One concern is that the LLM evaluator may have preferences related to response position, wording, length, or stylistic similarity to its own generations. We mitigate position effects with reversed-order evaluation and a conservative tie rule, but these measures do not eliminate all possible biases.

Another limitation is that all judgments come from a single evaluator model. Double-order evaluation measures its preference stability, but it does not establish agreement with other LLMs or human annotators. Future work should involve multiple evaluator models and human assessment.

A third issue is that the hosted evaluator model may be updated without a change to the public identifier. We report the model identifier, provider, evaluation date, prompt, sampling seed, and decoding configuration to support reproducibility.

Finally, LLM-based judgments complement automatic and human evaluation rather than replace either. Our claims are therefore limited to the comparison between DCC and our reproduced CEM under this specific protocol.

\section{Human Evaluation}
\label{sec:human_evaluation}

We complement the automatic and LLM-based evaluations with a
small-scale human preference study comparing DCC with the reproduced
CEM baseline.

\subsection{Evaluation Protocol}

We first exclude the 86 test instances for which DCC and CEM produce
identical responses, since these pairs provide no preference
information. We then uniformly sample 100 instances from the remaining
5,169 cases. For each instance, annotators are shown the situation,
dialogue context, and two anonymous responses. The reference response
is not provided.

The responses are presented as Response A and Response B, with DCC
assigned to each position exactly 50 times. Three annotators evaluate
all 100 pairs independently without access to model identities. For
each pair, they select Response A, Response B, or Tie according to the
following criteria:

\begin{itemize}
    \item \textbf{Empathy}: whether the response recognizes and
    appropriately addresses the speaker's feelings;
    \item \textbf{Relevance}: whether the response is consistent with
    the situation and dialogue context;
    \item \textbf{Overall quality}: the overall preference considering
    empathy, relevance, fluency, and usefulness.
\end{itemize}

A majority decision requires at least two matching judgments. Cases
in which the three annotators respectively select DCC, CEM, and Tie
are conservatively counted as ties. We use a two-sided exact binomial
test on the DCC and CEM majority decisions after excluding ties.
Inter-annotator agreement is measured using Fleiss' $\kappa$ ~\cite{fleiss1973equivalence} over the
three labels.

\subsection{Results}

Table~\ref{tab:human_evaluation} reports the majority-vote results.
DCC receives more preferences than CEM on all three criteria. The
largest difference occurs for empathy, where DCC is preferred in
48\% of the instances, compared with 37\% for CEM. DCC is also
preferred more often for relevance (50\% vs.\ 43\%) and overall
quality (42\% vs.\ 39\%).

\begin{table}[t]
\centering
\caption{Human preference evaluation on 100 test instances. Values
for DCC, CEM, and Tie are percentages based on majority voting.
The $p$-values are from two-sided exact binomial tests after excluding
ties.}
\label{tab:human_evaluation}
\setlength{\tabcolsep}{3.5pt}
\renewcommand{\arraystretch}{1.10}
\small
\begin{tabular}{lccccc}
\toprule
\textbf{Criterion}
& \textbf{DCC}
& \textbf{CEM}
& \textbf{Tie}
& \textbf{$p$}
& \textbf{$\kappa$} \\
\midrule
Empathy         & 48 & 37 & 15 & 0.278 & 0.064 \\
Relevance       & 50 & 43 &  7 & 0.534 & 0.040 \\
Overall quality & 42 & 39 & 19 & 0.824 & $-0.013$ \\
\bottomrule
\end{tabular}
\end{table}

After ties are removed, DCC accounts for 56.5\%, 53.8\%, and 51.9\%
of the decisive preferences for empathy, relevance, and overall
quality, respectively. However, none of the differences reaches
statistical significance. Fleiss' $\kappa$ is also relatively low across
the three criteria, reflecting the inherent subjectivity of human
preference judgments in open-ended dialogue generation. The human
evaluation therefore suggests a consistent preference toward DCC,
particularly for empathy, but should be interpreted as supplementary
rather than conclusive evidence.

\section{Computational Efficiency}
\label{sec:efficiency_analysis}

ICAD retrieves commonsense memories at every decoding step. To assess whether this imposes significant inference overhead, we measure the decoding efficiency of DCC and CEM on the same hardware. No additional training is required; we simply load the trained checkpoints and run greedy decoding on a single NVIDIA GeForce RTX 4090 GPU.

We use an effective generation batch size of 1 and a maximum decoding length of 50. Data-loading time is excluded; we measure only the execution time of the greedy generation function. Each run includes 20 warm-up batches followed by evaluation on 500 test instances, and we repeat the complete measurement three times. CUDA synchronization is performed before and after each generation call. Peak GPU memory is recorded via PyTorch's maximum allocated-memory statistic.

\begin{table}[t]
\centering
\caption{Efficiency comparison between CEM and DCC under greedy
decoding on the same NVIDIA GeForce RTX 4090 GPU. Latency is reported
as the mean and standard deviation over three runs of 500 test
instances after 20 warm-up batches. The effective generation batch
size is 1.}
\label{tab:efficiency_comparison}
\setlength{\tabcolsep}{3.8pt}
\renewcommand{\arraystretch}{1.10}
\small
\begin{tabular}{lrrrr}
\toprule
\textbf{Model}
& \textbf{Params}
& \textbf{Latency}
& \textbf{Samples/s}
& \textbf{Memory} \\
& \textbf{(M)}
& \textbf{(ms/sample)}
& 
& \textbf{(MB)} \\
\midrule
CEM
& 17.49
& $83.81 \pm 0.73$
& 11.93
& 91.82 \\
DCC
& 19.30
& $111.10 \pm 2.12$
& 9.00
& 98.78 \\
\bottomrule
\end{tabular}
\end{table}

As shown in Table~\ref{tab:efficiency_comparison}, DCC increases the
number of parameters from 17.49M to 19.30M, corresponding to a relative
increase of 10.32\%. Peak GPU memory increases by only 6.96MB, or
7.58\%. The small memory increase is consistent with the compact relation-level memories used by AGCF and ICAD.

DCC requires 111.10ms per response, compared with 83.81ms for CEM,
resulting in a latency increase of 32.55\%. Its throughput correspondingly
decreases from 11.93 to 9.00 samples per second. The median latency
increases from 74.71ms to 100.78ms, while the 95th-percentile latency
increases from 111.98ms to 138.56ms.

The extra computation is noticeable but limited. ICAD retrieves from two relation-level memories, each with only five COMET relation vectors--so each retrieval attends over just ten representations, not the full token-level commonsense sequences. Even though this happens at every decoding step, DCC still achieves roughly nine responses per second on a single RTX 4090.

Overall, dynamic commonsense coordination improves generation performance with modest increases in parameters and memory, at the expense of decoding latency. 
Since this comparison evaluates the complete DCC framework against CEM, the measured latency difference reflects the combined cost of SCE-AttnRes, AGCF, and ICAD and should not be interpreted as the isolated cost of ICAD alone.

\end{document}